\pdfoutput=1

\documentclass[11pt]{article}

\usepackage[]{acl}

\usepackage{times}
\usepackage{latexsym}

\usepackage[T1]{fontenc}

\usepackage{times}
\usepackage{latexsym}
\usepackage[T1]{fontenc}
\usepackage[utf8]{inputenc}
\usepackage{microtype}
\usepackage{amsmath,amsthm,amssymb}
\usepackage{enumitem}
\usepackage{algorithmic}
\usepackage{mathtools}
\usepackage{multirow}
\usepackage{booktabs}
\usepackage{graphicx}
\usepackage{caption}
\usepackage{subcaption}
\usepackage{float}
\usepackage{dsfont}
\usepackage{array}
\usepackage[ruled, lined, linesnumbered, commentsnumbered, longend]{algorithm2e}
\usepackage{fdsymbol}
\usepackage{url}

\usepackage{pifont}
\newcommand{\cmark}{\ding{51}}%
\newcommand{\xmark}{\ding{55}}%

\mathchardef\mhyphen="2D 

\newcommand{\method}{\textsc{FaiRR}}
\newcommand{\methodsmall}{\textsc{FaiRR}}
\newcommand{\baselinea}{ProofWriter}
\newcommand{\baselineaa}{PW (``All'')}
\newcommand{\baselineai}{PW (``Iter'')}
\newcommand{\baselineaaf}{ProofWriter (``All'')}
\newcommand{\baselineaif}{ProofWriter (``Iter'')}
\newcommand{\baselineb}{PRover}
\newcommand{\baselinebmulti}{multiPRover}

\DeclareMathAlphabet{\mathcal}{OMS}{cmsy}{m}{n}
\SetMathAlphabet{\mathcal}{bold}{OMS}{cmsy}{b}{n}

\newcolumntype{P}[1]{>{\centering\arraybackslash}p{#1}}

\title{FaiRR: Faithful and Robust Deductive Reasoning over Natural Language}

\author{ 
Soumya Sanyal\textsuperscript{{$1$}} \quad
Harman Singh\textsuperscript{{$2$}} \quad 
Xiang Ren\textsuperscript{{$1$}} \\
{{\textsuperscript{$1$}University of Southern California}} \quad {\textsuperscript{$2$}Indian Institute of Technology, Delhi} \\
{\texttt{\{soumyasa, xiangren\}@usc.edu, harmansingh.iitd@gmail.com}}
} 

\begin{document}
\maketitle

\begin{abstract}
	Transformers have been shown to be able to perform deductive reasoning on a logical rulebase containing rules and statements written in natural language. Recent works show that such models can also produce the reasoning steps (i.e., the \textit{proof graph}) that emulate the model's logical reasoning process. Currently, these black-box models generate both the proof graph and intermediate inferences within the same model and thus may be unfaithful.
	In this work, we frame the deductive logical reasoning task by defining three modular components: rule selection, fact selection, and knowledge composition. The rule and fact selection steps select the candidate rule and facts to be used and then the knowledge composition combines them to generate new inferences. This ensures model faithfulness by assured causal relation from the proof step to the inference reasoning.
	To test our framework, we propose \method{} (\underline{Fai}thful and \underline{R}obust \underline{R}easoner) where the above three components are independently modeled by transformers.
	We observe that \method{} is robust to novel language perturbations, and is faster at inference than previous works on existing reasoning datasets. Additionally, in contrast to black-box generative models, the errors made by \method{} are more interpretable due to the modular approach. \footnote{The source code of \method{} has been made available at \url{https://github.com/INK-USC/FaiRR}.}
\end{abstract}
	
\section{Introduction}
\label{sec:intro}

\begin{figure}[t]
	\centering
	\includegraphics[width=0.97\columnwidth]{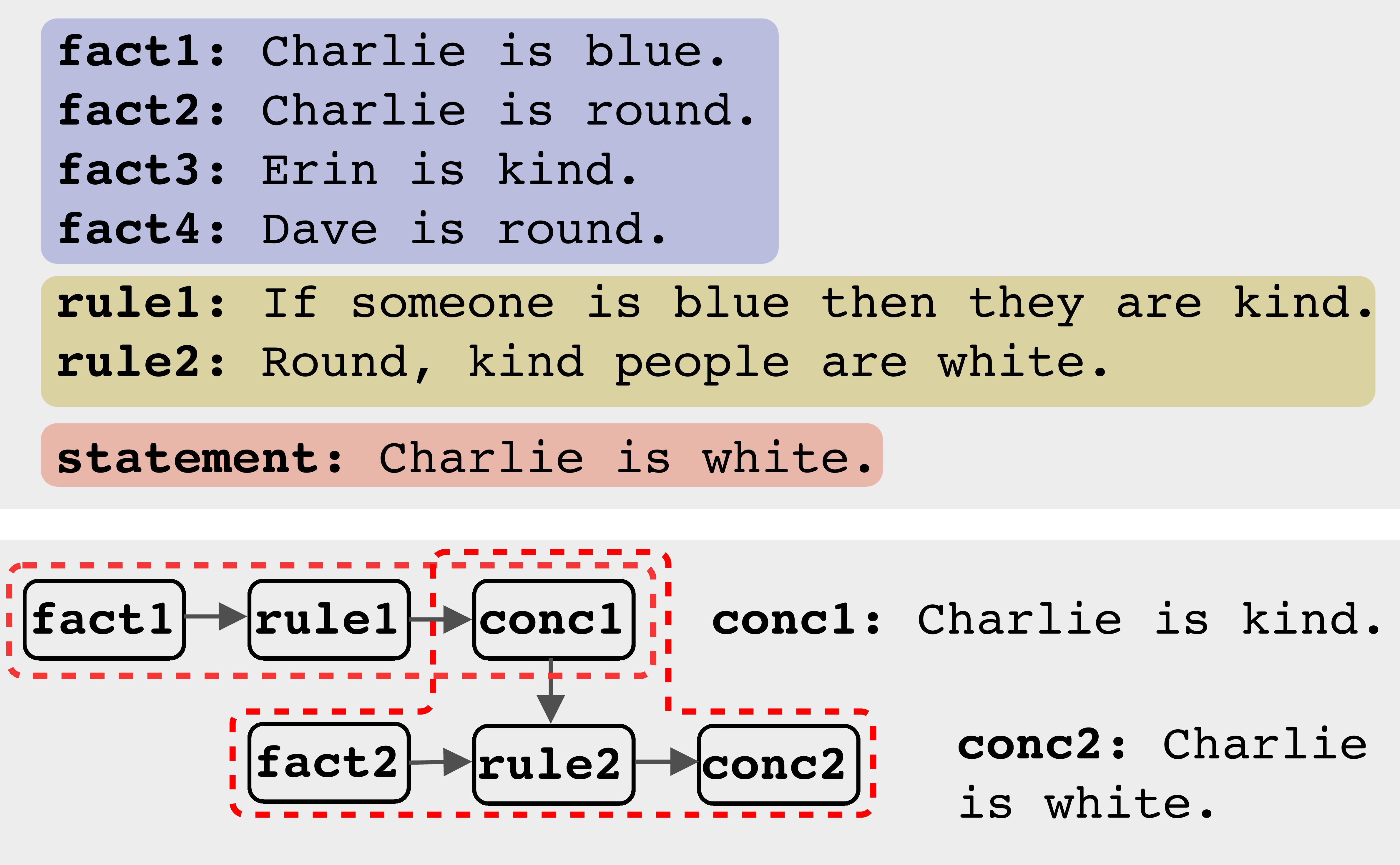}
	\caption{\textbf{Example of a theory, a statement, and a valid proof graph} - An instance contains multiple facts and rules in blue and yellow respectively, followed by a statement in red. The proof graph describes the reasoning steps required to generate the statement.}
	\label{fig:example}
\end{figure}

The field of AI has long pursued the goal of building systems that can automatically reason over some given \textit{explicit} knowledge to generate conclusions and provide the reasoning steps involved in the process \cite{Mccarthy1959ProgramsWC,Newell1956TheLT}.
Recently, \citet{ruletaker} proposed a modern version of this problem, where the formal representation of knowledge is replaced by natural language statements in English. Further, they proposed a transformer-based model \cite{vaswani2017attention} RuleTaker, that can predict if a candidate statement is entailed by the natural language statements, by emulating deductive reasoning.
As shown in Figure \ref{fig:example}, in this deductive reasoning task, facts and rules from the rulebase are combined iteratively to generate intermediate inferences which eventually entails the statement. Note that the reasoning process implicitly involves two steps: determining which rules and facts to combine at each iteration, followed by using them to generate an intermediate conclusion.

While RuleTaker focuses on just predicting the statement entailment, some recent works \cite{prover,proofwriter} have further developed systems that can also generate the reasoning steps (i.e., \textit{proof graph generation}). However, these systems do not \textit{explicitly} ensure the causality from the rule/fact selection to generating the intermediate inferences. Since these systems are inherently black-box models, it is unclear if such constraints are implicitly learned by the models without being enforced externally. This, in turn, questions the faithfulness of the model's internal reasoning process \cite{lipton2018themythos}. Because the model has access to the full theory at input, it might use additional parts of the theory, than just the predicted proof, to generate the inference.

\begin{figure}[t]
	\centering
	\includegraphics[width=0.9\columnwidth]{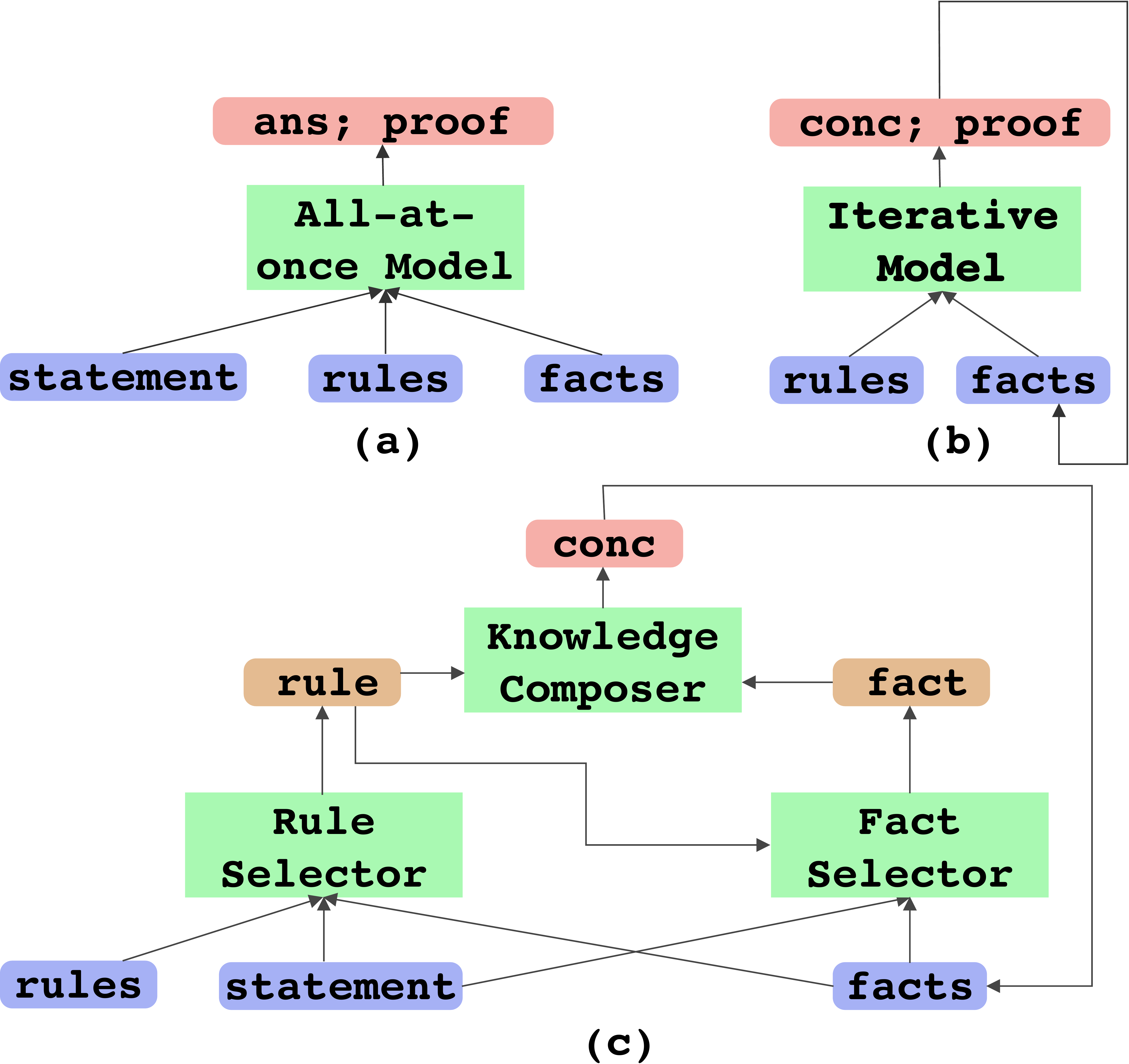}
	\caption{\textbf{Reasoning process in different models}. \textbf{(a)}: \baselineaaf{} directly output the entailment prediction and proof graph for given input. \textbf{(b)}: \baselineaif{} iteratively generates the one-step intermediate conclusions and their proofs. \textbf{(c)}: \method{} selects a rule, then a fact, and finally combines them to generate an intermediate inference. Note that the proof is implicitly determined by the selection steps. Please refer to Section \ref{sec:method_causal} for details.}
	\label{fig:causal}
\end{figure}

In this paper, we address these shortcomings by developing a modularized framework to solve the deductive reasoning task. While existing methods generate both proofs and conclusions in a single step, in our framework we break this process into three steps: rule selection, fact selection, and knowledge composition. The rule selection step decides the relevant rule to use for an iterative inference step and fact selection uses this rule to select the relevant facts. Then, the knowledge composition step reasons using only the selected rule and facts to generate the next intermediate inference. In Figure \ref{fig:causal}, we show the model schematics for our system and contrast it with previous methods. Notably, we strictly restrict the information accessible at each step of our framework to make the reasoning process more faithful. For example, the fact selection step depends only on the selected rule, instead of all the rules in the rulebase. Additionally, the generated inference depends  \textit{explicitly} on the selected rule and facts, as opposed to all the rules and facts in prior works. This makes the proof graph a \textit{by-product} of the selection steps as we don't need to generate any separate proofs. Since we constrain the inputs to each step, this also makes each sub-problem easier to learn, leading to an overall more robust reasoning model.

To model these three steps, we develop \method{}, in which each component is a transformer-based model learning to perform the modular tasks. Specifically, we use RoBERTa-based models \cite{liu2019roberta} for the two selection tasks and a T5-based model \cite{raffel2019exploring} for the composition task. Similar to \baselinea{}, we use synthetic rulebases to train \method{}. To test the deductive reasoning capabilities in a more comprehensive way, we experiment with both existing deductive reasoning datasets and multiple newly-generated robustness dataset variants. Overall, we find that \method{} is more robust to novel language perturbations than baselines. Additionally, our model is up to three times faster at inference due to the constrained input and outputs of different modules. Lastly, we find that the errors made by our model are more interpretable and easier to debug compared to baseline generative models. This further demonstrates the faithfulness of our modularized reasoning framework.
\section{Problem Definition}
\label{sec:definition}
\vspace{-0.1cm}
\paragraph{Notations} 
A theory $T$ consists of a set of facts \begin{small}$F=\{f_1, f_2, \dots, f_n \}$\end{small} and rules \begin{small}$R=\{r_1, r_2, \dots, r_m\}$\end{small} expressed in natural language. 
An example of a theory is depicted in Figure \ref{fig:example}. Here, the sentences in the blue and yellow boxes are facts and rules, respectively. Further, a \textit{proof graph} is a directed graph connecting facts and rules that describe how a specific inference can be obtained from the theory. In Figure \ref{fig:example}, the proof graph shows the steps involved in generating the inference ``\textit{Charlie is white.}''. To generate the proof graph we may need to infer some intermediate conclusions $c_i$. These inferences are considered as part of the extended facts in the theory. For example, in Fig.~\ref{fig:example}, ``\textit{Charlie is kind}'' is an intermediate inference required to generate the correct proof graph.

\paragraph{Deductive Reasoning} The task of deductive reasoning is described as follows: given a theory $T$, and a statement $s$, predict if the theory supports the statement (\textit{entailment prediction}) and if so, generate the proof graph that supports the statement (\textit{proof generation}). For the example theory and statement in Figure \ref{fig:example}, we see that the statement is indeed entailed by the theory and the valid proof graph is shown for the same. The main goal of this task is to evaluate if a model can generate valid reasoning chains in the form of proof graphs to justify its entailment prediction.

\paragraph{Reasoning Robustness} We consider an auxiliary task that evaluates the robustness of the reasoning abilities used by the model. Let $\mathcal{P}$ be a perturbation function that modifies a given theory $T$ (statement $s$) to a theory $T'$ (statement $s'$), such that $(T', s')$ just has some surface changes in the natural language form but still requires the similar reasoning process as required for $(T, s)$. A function that alters the subjects in the theory to unseen subjects is an example of such perturbation function. We perturb each theory statement pair $(T, s)$ to create an \textit{equivalence set} defined as the set $E_{(T, s)} = \{(T'_1, s'_1) \dots (T'_N, s'_N)\}$, where each $(T'_k, s'_k)$ is derived by perturbing the original theory, and $N$ is the total such perturbations per theory. Note that it is possible to generate different $(T'_k, s'_k)$ pairs by controlling the stochasticity of $\mathcal{P}$. The main goal of this task is to evaluate the consistency of the model's  predictions with minimal variations in the input theory.

\paragraph{Evaluation Protocol}
We consider three main aspects for evaluating the model performance in our study:
(1) \textbf{\textit{Entailment accuracy}} measures how accurately the model is able to predict the true statement entailment.
(2) \textbf{\textit{Proof accuracy}} measures how accurately the model can predict a valid proof for the statement. Following \citet{prover,proofwriter}, we use the strict metric for proof evaluation, i.e., for a match to count, both the predicted proof should exactly match a gold proof and the entailment should be correctly predicted.
(3) \textbf{\textit{Consistency}} measures if the models are consistent in the entailment and proof prediction for different perturbation functions. For a theory statement pair $(T, s)$ and its corresponding equivalence set $E_{(T, s)}$, consistency is defined as $C = \frac{1}{N}\sum_{k=1}^{N} \mathds{1}[f(T, s) = f(T_k, s_k) ],$
where $f(\cdot)$ is the model's prediction. We compute the average consistency for both entailment and proof predictions on an equivalence set and further average across the dataset to report the consistency.

\section{The \method{} Method}

\subsection{Approach Overview}
\label{sec:method_causal}
As illustrated by the example in Figure \ref{fig:example}, to reliably generate a proof graph through deductive reasoning, a model needs to generate multiple one-hop intermediate conclusions. This is the major limitation of models that use the theory to directly predict the proof (Figure \ref{fig:causal} (a)), thus questioning the trustworthiness of the reasoning process. Next, it is also intuitive to see that in order to faithfully generate these intermediate inferences, a model should first determine the proof (i.e., know the rules and facts to use) and then use them to infer the conclusion. That is, there is a causal relation from determining the proof to then generating the conclusion. We note that \baselineaif{} lacks in this aspect. As shown in Figure \ref{fig:causal} (b), it first generates the conclusion and then the corresponding proof.

Motivated by these points, we propose our causal reasoning framework which breaks the reasoning process into three desirable steps. As shown in Figure \ref{fig:causal} (c), in our framework, first a rule $r$ is selected using the rules and facts in the theory. Following that, some relevant facts are selected from the fact list based on the selected rule $r$. This step does not use the other rules $R \setminus \{r\} $ in the theory. Finally, the selected rule and facts are jointly used to generate a new conclusion $c_i$. In this framework, the one-step proof is explicitly determined first via the selection steps followed by the inference generation, making the proof a \textit{by-product} of the whole process. In contrast, prior works learned to generate the proof along with intermediate conclusion.

\subsection{\method{} Modules}
\label{sec:method_modules}
At a high level,
\method{} is an iterative model in which the one-hop intermediate conclusions are generated step-by-step. To model our framework described in Sec.~\ref{sec:method_causal}, we have four components in \method{} as follows.

\begin{figure}[t]
	\centering
	\includegraphics[width=0.8\columnwidth]{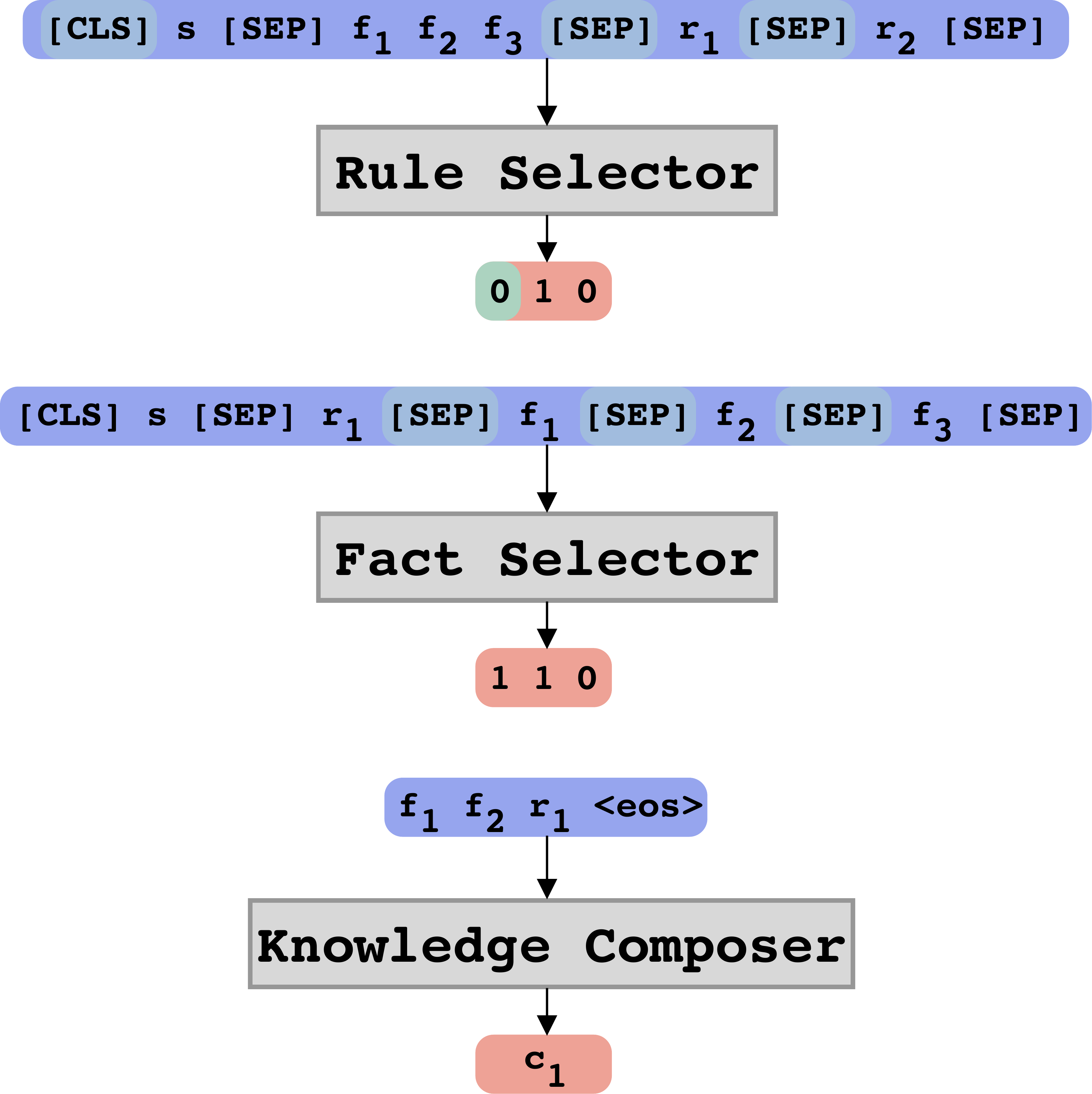}
	\caption{ \textbf{Overview of components of \method{}} - The rule selector and fact selectors are classification models whereas the knowledge composer is a generation model. The input tokens used for classification by the selectors are highlighted. Rule selector decides to stop based on the output prediction of [CLS] token (highlighted in green). Here, rule $r_1$, and facts $f_1$ and $f_2$ are used to generate the conclusion $c_1$. Please refer to Section \ref{sec:method_modules} for more details.}
	\label{fig:model}
\end{figure}

\paragraph{Rule Selector (RS)}
	The rule selector is a RoBERTa-based \cite{liu2019roberta} classification model that takes the concatenated statement, facts, and rules as input, and selects a rule that is used to generate an intermediate conclusion in the current iterative step. It takes the input of the form \begin{small}$[CLS] ~s~ [SEP] ~F~ ~[[SEP]~ ~r_i~]_m ~[SEP]~$\end{small},
	and generates a one-hot output vector by classifying the token embedding from the [CLS] token and [SEP] tokens in front of the rules, via a linear classifier layer. Each classification is a binary classification, but overall only one of the tokens has the positive class. Here $s$ denotes the statement, $F$ is the facts and concatenated with any intermediate conclusions generated in a prior iteration, and $\{r_i\}$ denotes the $i^{th}$ rule in the theory that contains a total of $m$ rules. $[~]_m$ denotes continued concatenation. An example input and output of the rule selector is shown in Figure \ref{fig:model}. If a [SEP] token is selected, we select the rule sentence following the corresponding [SEP] token, otherwise if the [CLS] token is selected, we decide to stop the iteration. That is, the [CLS] selection acts as a stop signal for our iterative model. We note that it is possible to have more than one likely candidate rule since there can be multiple one-hop inferences possible for a given theory. Following \citet{proofwriter}, we randomly select one of the possible candidate rules at each iteration.

\paragraph{Fact Selector (FS)}
	The fact selector is RoBERTa-based \cite{liu2019roberta} token classification model that takes the statement, the rule selected by the rule selector, and facts in the theory, and then predicts a set of candidate facts that can be used with the rule to generate an intermediate conclusion. It takes the input of the form \begin{small}$[CLS] ~s~ [SEP] ~r~ ~[[SEP]~ ~f_i]_n~ ~[SEP]~$\end{small},
	where $s$ is the statement, $r$ is the selected rule, and $\{f_i\}$ is the $i^{th}$ fact in the theory containing $n$ total facts. Note that facts also include any previously generated intermediate conclusions. $[~]_n$ denotes continued concatenation. The output is generated by classifying each [SEP] token embedding in front of a fact using a linear layer, to determine if the corresponding fact is selected or not. An example input and output for the fact selector is depicted in Figure \ref{fig:model}. We note that it is possible to have some rules that can reason over multiple facts jointly to generate a conclusion. An example of such a rule is ``\textit{rule2}'' in Figure \ref{fig:example}. Hence, this component has the ability to select multiple facts.

	\paragraph{Knowledge Composer (KC)}
	The knowledge composer is a generative text-to-text transformer T5 \cite{raffel2019exploring} (T5-large) that can compose a set of facts and a rule to output a novel conclusion. The input to the model is the selected facts and rule concatenated together, and the output is the intermediate conclusion. An example input and output for knowledge composer is shown in Fig.~\ref{fig:model}.

\paragraph{Solver}
	The final component is the solver that operates after all iterations have finished (i.e., once the rule selector selects the [CLS] token indicating to stop the iterative inference generation process). Similar to \baselinea{}, our solver currently searches for the statement in the generated intermediate inferences (string matching). If found, it predicts that the statement is entailed by the theory. It also search for the negation of the statement \footnote{Following \baselinea{}, we perform regex to add/remove ``not'' which suffices for this dataset.}, and if found, it predicts not entailed. If none of these are present, it predicts ``Unknown'' since it cannot prove or disprove the statement. The proof graph is constructed by using the one-hop proofs generated by the selected rule and facts at each step. For example, in Figure \ref{fig:example}, the red dotted boxes (one-hop proofs) are stitched together to assemble the complete proof. For cases where the entailment is ``Unknown'', the proof returned is ``None'', since no proof for the statement exists in the theory. We note that our solver is not a learnable module.

\subsection{Training and Inference}
Each component of our model (except the solver, which is deterministic) is trained separately. We use the same dataset as \baselinea{} to train these models, but process it such that each model receives only the relevant inputs according to our causal framework. More concretely, suppose for a given theory $T = R + F$, a possible intermediate inference is $c$ obtained by using a rule $r$ and a fact $f$. Then, a training instance of \baselinea{}, which is a T5 \cite{raffel2019exploring} model, uses the input $\{R, F\}$ and output $\{c, r, f\}$. We process the same instance to generate three training instances, one for each of rule selector, fact selector, and knowledge composer, respectively, as follows:
\begin{align*}
RS~Input = \{R, F\}; ~~RS~Output = \{r\},\\
FS~Input = \{r, F\}; ~~FS~Output = \{f\},\\
KC~Input = \{r, f\}; ~~KC~Output = \{c\}.
\end{align*}
Our selector models have the statement $s$ as input to the model. Also, the outputs of rule selector and fact selectors are converted to class labels instead of text since our selectors are classification models. We use cross entropy loss to train the rule selector, and binary cross entropy loss to train the fact selector. The knowledge composer is trained on language modeling loss.

At inference time, the rule selector selects a rule to be used for generating one-step conclusions. Then, the fact selector selects some facts based on the selected rule, which is then collectively passed on to the knowledge composer to generate a conclusion. This three-step pipeline is run iteratively until the rule selector predicts a stop signal by selecting the [CLS] token which exits the iteration. Once the iteration finishes, the solver uses the generated intermediate inferences to decide if the statement is entailed or not, and generates a proof accordingly.

\paragraph{Remark on Computational Complexity}
\label{sec:method_computation}
A practical limitation of \baselinea{} is that it performs an exhaustive forward search by enumerating all possible inferences from a given theory. This leads to redundant inferences being generated for proving a particular statement. Additionally, using a text-to-text transformer model adds to the problem since it is usually quite expensive to run at inference time. In \method{}, we alleviate this by introducing two changes. First, our causal framework allows only selected rule and facts as input to the knowledge composer, thus restricting the input length significantly. Second, augmenting the question to our selector inputs helps reduce the candidate space because these models can learn to prioritize the selection based on the relevance to both the question and the theory. This ensures that \method{} does not perform an exhaustive forward search and prioritizes generating relevant inferences over the others. Both these changes lead to an overall improvement in inference speed. We perform more quantitative analysis on this later in Section \ref{sec:res_time_analysis}.
\section{Experimental Setup}
\label{sec:setup}

\paragraph{Datasets}
Following \cite{proofwriter,ruletaker}, we use the D* datasets for our experiments. These are a set of multiple datasets - namely D0, D1, D2, D3, D0-D3, and D5. The theory in these datasets are synthetically generated with increasing reasoning depths. For example, D3 dataset contains statements that require \textit{at most} 3-hop reasoning steps. The D0-D3 contains all theories in D3 plus $\sim 20\%$ of the D0-D2 training set theories. We also use the ParaRules dataset \cite{ruletaker} that contains around 2k theories expressed in paraphrased natural language. 

Additionally, we generate three datasets that evaluate the robustness of the reasoning models as follows:
\begin{itemize}[noitemsep,topsep=4pt]
	\item \textbf{\textit{Subject} robustness}: Here, subjects in a theory are perturbed by using some out-of-distribution proper and common names. For example, in Figure \ref{fig:example}, ``\textit{Charlie}'' can be replaced with ``\textit{Paul}'' which is not used in the D* datasets. We generate five new theories corresponding to each theory of the D3 dataset, by repeatedly perturbing all the proper and common names in the theory.
	\item \textbf{\textit{Attribute} robustness}: Here we sample out-of-distribution attributes. For example, ``\textit{blue}'' in Figure \ref{fig:example} can be replaced with ``\textit{soft}''. As above, we generate five new theories for each theory of the D3 dataset.
	\item \textbf{\textit{Subject+Attribute} robustness}: This is a combination of subject and attribute robustness to study model performance when most of the training vocabulary is replaced by out-of-distribution words. Each theory has both novel subject and attribute.
\end{itemize}
We include more details on the perturbation sets used in our experiments in Appendix \ref{app:robustness_dataset}.

\paragraph{Baselines}  
We compare \method{} with two variants of \baselinea{} \cite{proofwriter}: All-at-once (\baselineaa{}) and Iterative (\baselineai{}), wherever applicable \footnote{\scriptsize The code to reproduce numbers of \baselinea{} is not publicly available. We either copy results directly from the paper or run our own inference on model checkpoints made available by the authors.}. The \baselineaa{} model is trained to predict the entailment and generate proof graph directly from the theory and statement in a single step. The \baselineai{} generates one-step inferences and corresponding proofs iteratively, until all possible inferences are generated, and then stitches the proof graph similar to our method. If not mentioned otherwise, \baselinea{} uses a T5-large \cite{raffel2019exploring} model. We omit comparisons with \baselineb{} since it was trained on a different dataset that adds specific constraints on the proof graph. Please refer to Appendix \ref{app:comp_baselines} for more details.
\section{Experiment Results}
\label{sec:results}

We compare \method{} with \baselinea{} variants on three settings: generalization on D* datasets, robustness to perturbed theories, and efficiency in inference computation. We further conduct qualitative analysis to understand the inference errors.

\begin{table}[t]
	\centering
	\resizebox{0.88\columnwidth}{!}{%
		\begin{tabular}{lcccc}
			\toprule
			& \multicolumn{2}{c}{\textbf{Entailment Accuracy}}	& \multicolumn{2}{c}{\textbf{Proof Accuracy}}	\\
			\cmidrule(r){2-3} \cmidrule(r){4-5}
			$d$ & \baselineai{} & \methodsmall{} & \baselineai{} & \methodsmall{}	\\
			\midrule
			N/A & 99.7 & 99.6 & 99.7 & 99.6 \\
			0 & 100.0 & 100.0 & 100.0 & 100.0 \\
			1 & 99.9 & 99.7 & 99.9 & 99.5 \\
			2 & 99.7 & 98.9 & 99.4 & 97.2 \\
			3 & 99.7 & 96.6 & 99.1 & 95.3 \\
			\midrule
			All & 99.8 & 99.2 & 99.7 & 98.8 \\
			\bottomrule
		\end{tabular}%
	}
	\caption{\label{tab:results_d03_d03} Comparison of \method{} with \baselineaif{} trained and tested on D0-D3. Baseline results are generated using the checkpoint provided by the authors. For more details please refer to Section \ref{sec:res_depth}.}
\end{table}

\subsection{Performance on Same Depth Reasoning}
\label{sec:res_depth}
In this setting, we train and test both models on D0-D3 dataset. Note, D0-D3 contains statements with reasoning depths up to 3. This compares the ability of the models to generalize to \textit{seen} reasoning depths at train time. The results with increasing depths of reasoning are shown in Table \ref{tab:results_d03_d03}. Here, depth ``N/A'' refers to statements that cannot be proven and hence don't have an exact proof depth associated with it. We observe that overall both \method{} and \baselineaif{} performs comparably (last row with depth 'All'). Further, we find that our model's performance is lower on $d=3$, indicating that our models tend to perform weaker with increasing depths. This happens majorly because the rule selector in \method{} tends to incorrectly select the [CLS] token to indicate a stop signal instead of generating more possible intermediate inferences. We discuss more about this in Sections \ref{sec:res_time_analysis} and \ref{sec:res_error_analysis}. Please refer to Appendix \ref{app:depth_generalization} for more results on unseen reasoning depths.

\begin{table}[t]
	\centering
	\resizebox{0.94\columnwidth}{!}{%
		\begin{tabular}{lcccccc}
			\toprule
			\multirow{2}{*}{Robustness} & \multicolumn{3}{c}{\textbf{\baselineai{}}} & \multicolumn{3}{c}{\textbf{\methodsmall{}}} \\
			\cmidrule(r){2-4} \cmidrule(r){5-7}
			& EA & PA & C	& EA & PA & C \\
			\midrule
			Subject & 89.6 & 88.4 & 87.6 & 96.8 & 95.9 & 96.4 \\
			Attribute & 97.8 & 97.4 & 97.4 & 96.7 & 95.6 & 96.5 \\
			Subject+Attribute & 94.8 & 93.4 & 93.7 & 95.4 & 94.3 & 94.7 \\
			\midrule
			Average & 94.1 & 93.1 & 92.9 & 96.3 & 95.3 & 95.9 \\
			\bottomrule
		\end{tabular}%
	}
	\caption{\label{tab:results_robust} Comparison of \method{} with \baselineaif{} when trained on D0-D3 dataset and tested on different robustness datasets. EA, PA, and C refers to entailment accuracy, proof accuracy, and consistency, respectively. Please refer to Section \ref{sec:res_robust} for more details.}
\end{table}

\begin{table}[t]
	\centering
	\scalebox{0.7}{
		\begin{tabular}{lcccc}
			\toprule
			& \multicolumn{2}{c}{\textbf{Entailment Accuracy}}	& \multicolumn{2}{c}{\textbf{Proof Accuracy}}	\\
			\cmidrule(r){2-3} \cmidrule(r){4-5}
			$d$ & \baselineai{} & \methodsmall{} & \baselineai{} & \methodsmall{}	\\
			\midrule
			N/A & 98.9 & 99.3 & 98.9 & 99.3 \\
			0 & 99.9 & 100.0 & 99.9 & 100.0 \\
			1 & 79.1 & 96.0 & 78.8 & 95.7 \\
			2 & 76.6 & 93.4 & 73.4 & 91.4 \\
			3 & 72.7 & 89.8 & 67.8 & 85.7 \\
			\midrule
			All & 89.6 & 96.8 & 88.4 & 95.9 \\
			\bottomrule
		\end{tabular}%
	}
	\vspace{-0.2cm}
	\caption{\label{tab:results_d03_subject_depth_new} Comparison of \method{} with \baselineaif{} trained on D0-D3 and tested on subject robustness dataset. Baseline results are generated using the checkpoint provided by the authors. For more details please refer to Section \ref{sec:res_robust}.
	}
\end{table}

\subsection{Robustness to Perturbed Theories}
\label{sec:res_robust}
In this section, we test the robustness of \baselineaif{} and \method{} on different perturbed theories. Since \method{} focuses on making deductive reasoning more robust and faithful, performance on these robustness experiments are the main results of our work.
As described in Section \ref{sec:setup}, we test the robustness on three different perturbations: subject, attribute, and subject+attribute. We compare the performance of both models after training on D0-D3 dataset. The consolidated results are shown in Table \ref{tab:results_robust} and depth-wise results for subject robustness are shown in Table \ref{tab:results_d03_subject_depth_new}. We report the entailment accuracy, proof accuracy, and consistency as defined in Section \ref{sec:definition}. Please refer to appendix \ref{app:robustness_depth} for the depth-wise breakdown of all the datasets. We observe that on subject and subject+attribute robustness, our models are consistently better than \baselinea{} whereas on attribute robustness both models perform similarly. Further, we find that on average, \method{} is both more accurate and consistent than the baseline.
From this, we conclude that our model relies less on spurious correlations based on the subject while both models likely suffer from similar issues on attribute perturbations. Since \baselinea{} uses the theory to \textit{generate} the intermediate conclusion and proofs, it has the capacity to exploit some spurious patterns that can inflate performance. In contrast, our causal framework restricts this capacity by constraining the inputs to each component as described in Section \ref{sec:method_causal}. Hence, these robustness evaluations demonstrate one of the prime benefits of our causal and modular approach.

\subsection{Study on Inference Efficiency}
\label{sec:res_time_analysis}
Here we perform several analyses to evaluate the computational benefits of our method as described in Section \ref{sec:method_computation}. Inference efficiency is an important aspect of this problem for real-world scenarios where compute can be limited.

\begin{figure}
	\centering
	\begin{subfigure}{.25\textwidth}
		\centering
		\includegraphics[width=\columnwidth]{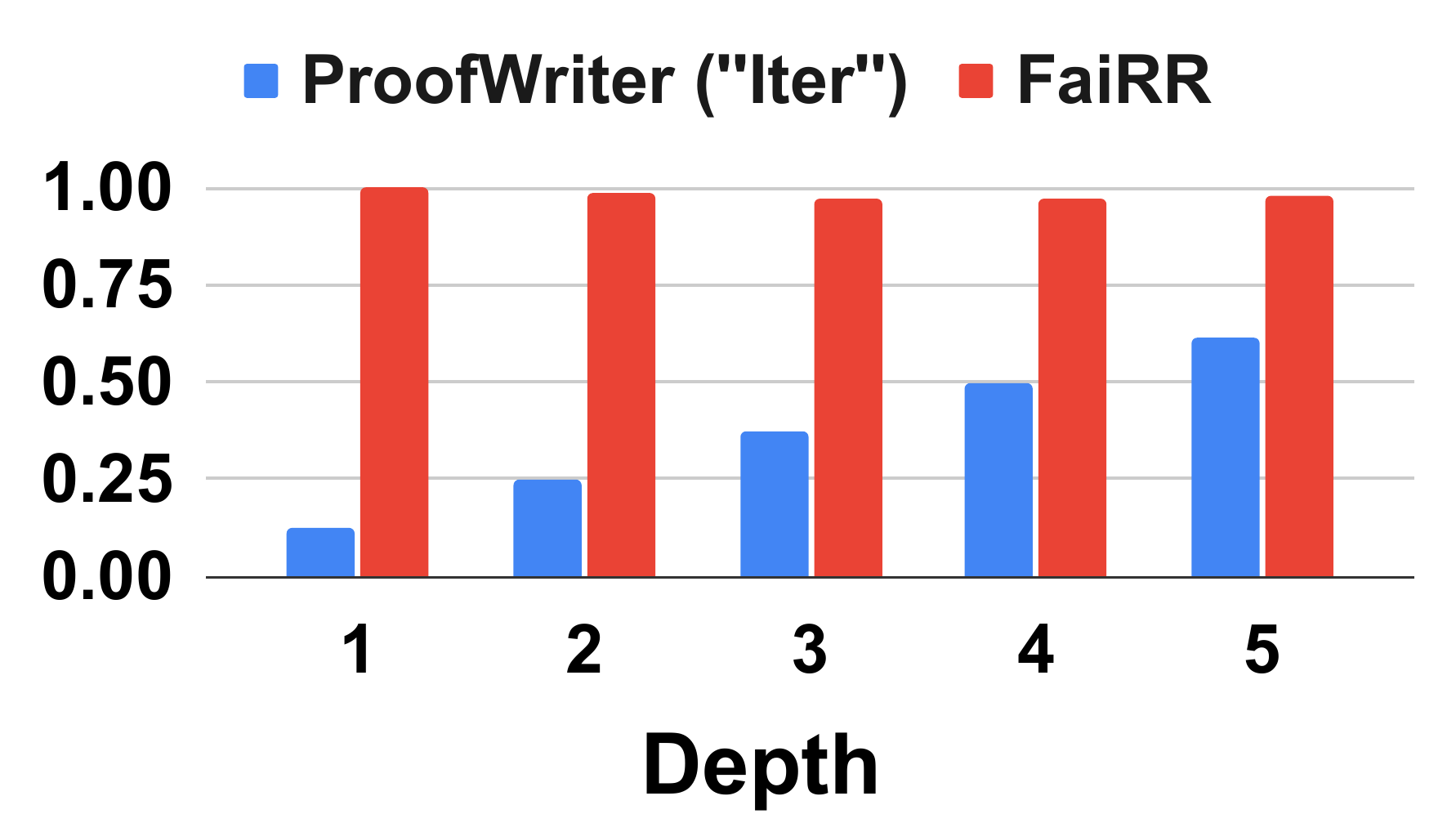}
		\caption{Precision}
	\end{subfigure}%
	\begin{subfigure}{.25\textwidth}
		\centering
		\includegraphics[width=\columnwidth]{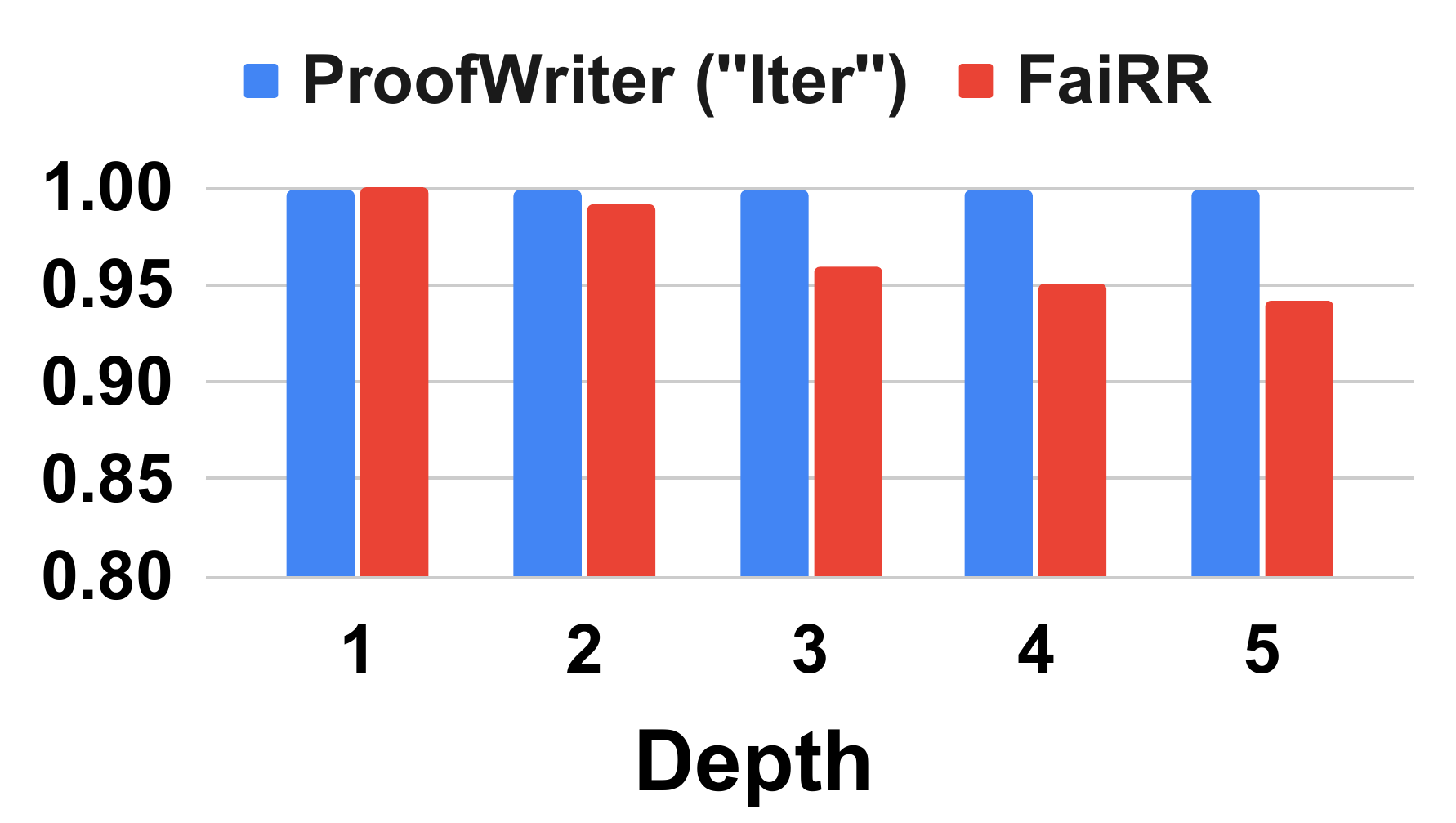}
		\caption{Recall}
	\end{subfigure}%
	\caption{\label{fig:pr_curve} Comparison of \baselineaif{} and \method{} on precision and recall of generated inferences with increasing reasoning depths.
	}
\end{figure}

\paragraph{Relevance of generated inferences} Here, we study the relevance of the intermediate inferences generated by \method{} and \baselineaif{}. Let $T$ be the set of intermediate inferences required for generating the proof graph for the statement.
Further, let $G$ be the set of intermediate inferences actually generated by a model. Then, the precision and recall are defined as $P = \frac{|T \cap G|}{|G|},~\textrm{and}~ R = \frac{|T \cap G|}{|T|}$
In Figure \ref{fig:pr_curve}, we plot the precision and recall for both \method{} and \baselineaif{} with increasing reasoning depths. We find that our model has close to $1.0$ precision at all depths, whereas \baselinea{} has low precision. This demonstrates that our model is able to successfully prune the candidate inference space to generate relevant candidate inferences almost perfectly. In contrast, we see that with increasing depths, our model's recall reduces from close to $1.0$ to $\approx 0.95$ whereas \baselinea{} has a perfect recall at all depths. While the drop is not very drastic, it indicates that our model fails to generate some essential inferences at higher depths. This is mainly because our rule selector decides to stop early and not generate further relevant inferences for some provable statements. Overall, we conclude that \method{} always generates inferences that are relevant to solving the instance, although at higher depths it can miss some relevant conclusions.

\begin{figure}[t]
	\centering
	\begin{subfigure}{.5\columnwidth}
		\centering
		\includegraphics[width=\columnwidth]{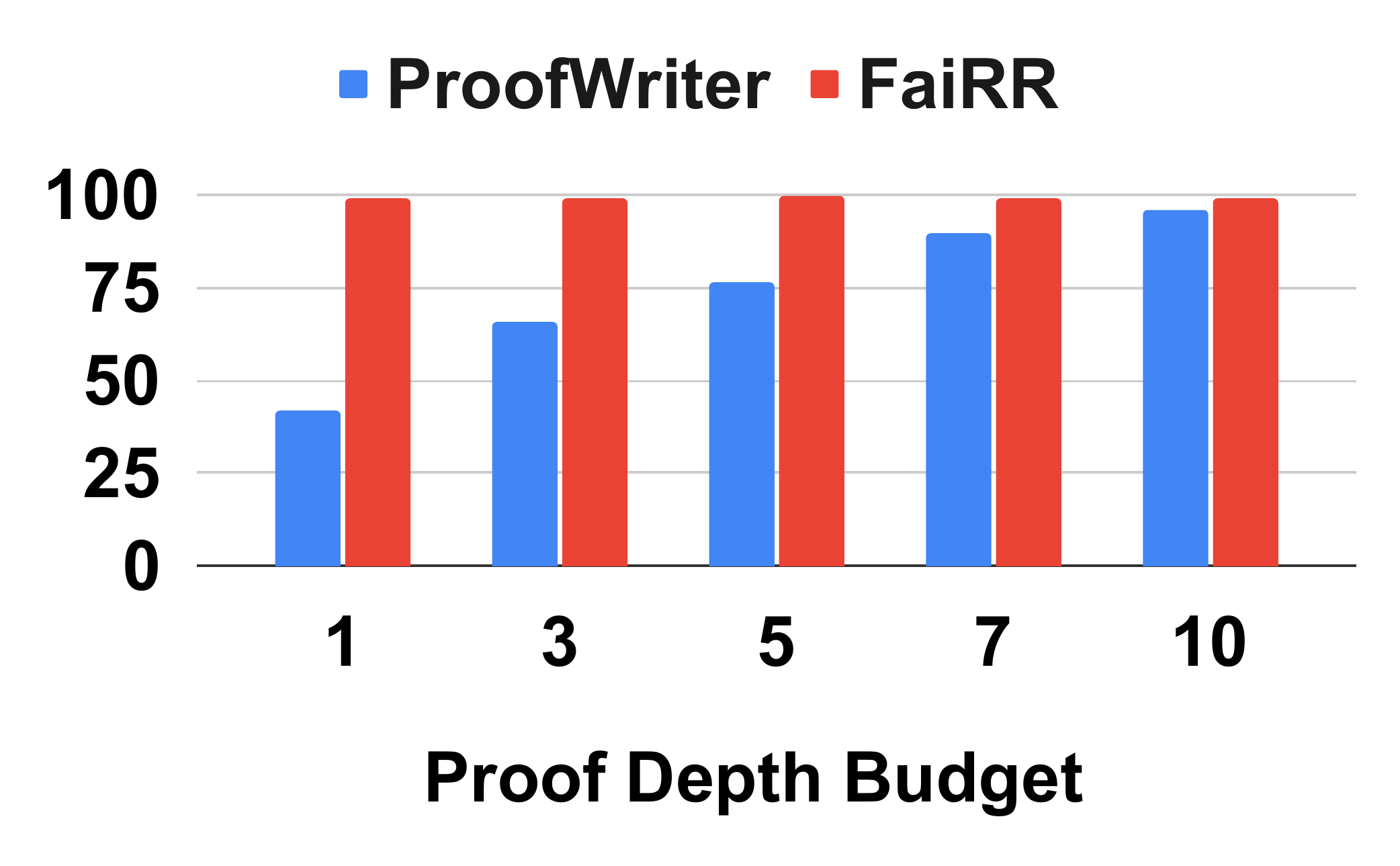}
		\caption{Depth 1}
	\end{subfigure}%
	\begin{subfigure}{.5\columnwidth}
		\centering
		\includegraphics[width=\columnwidth]{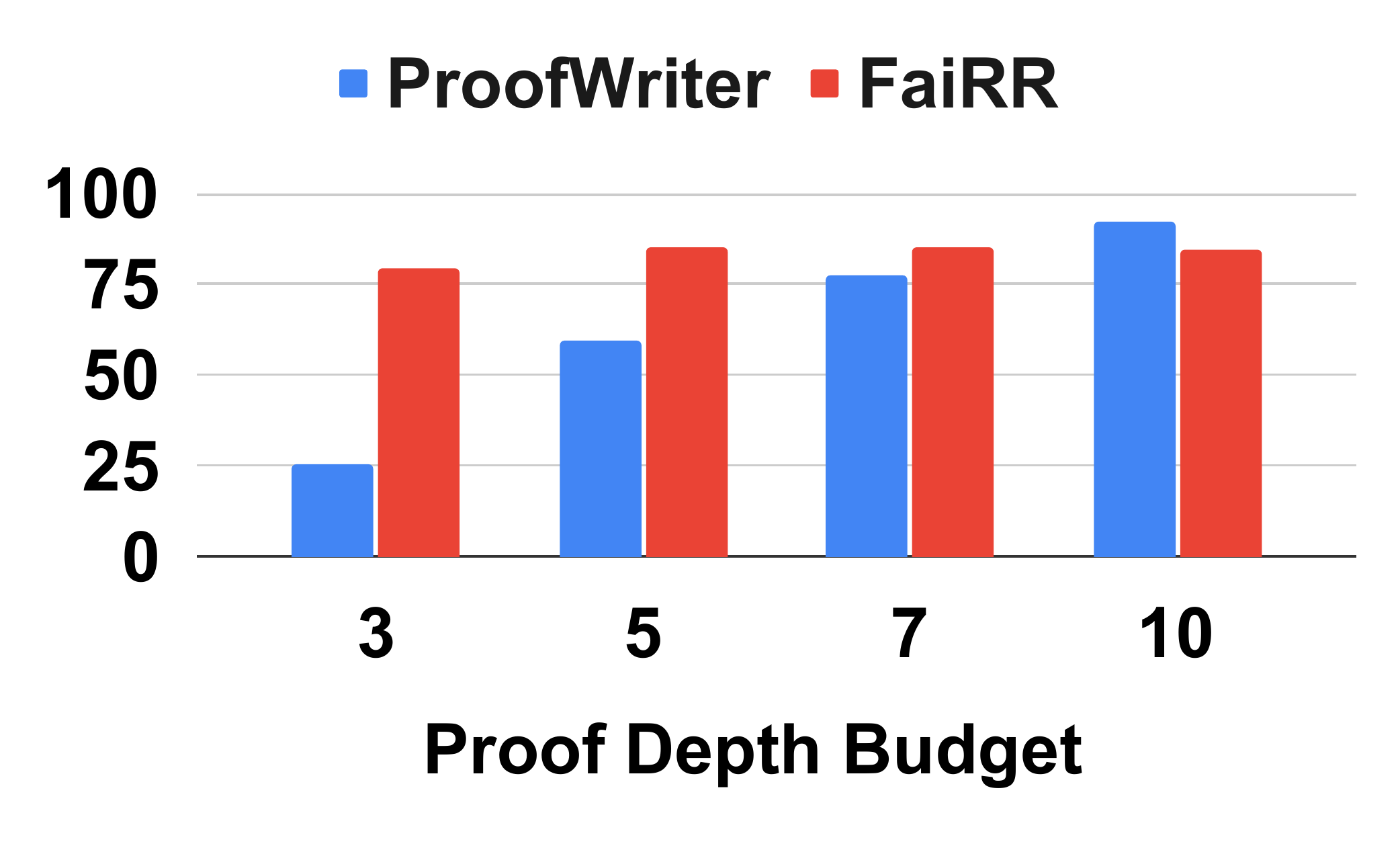}
		\caption{Depth 3}
	\end{subfigure}%
	\caption{\label{fig:budget_small} Depth-wise comparison of \baselineaif{} and \method{} on limited inference budgets. Please refer to Section \ref{sec:res_time_analysis} for details.
	}
\end{figure}

\paragraph{Performance under inference budget constraints}
We analyze the performance of \method{} and \baselinea{} under a fixed inference budget constraint by restricting the total number of conclusions that can be generated. We perform this analysis for different reasoning depths and depict the results in Figure \ref{fig:budget_small}. We observe that \method{} consistently outperforms \baselinea{} on lower budgets. This shows that \method{} performs a prioritized generation of conclusions that are relevant to the statement, which can be useful in scenarios with limited inference budgets. See Appendix \ref{app:budget_analysis} for more comparisons.

\paragraph{Inference runtime analysis} We next compare the time taken by both the models to solve the complete D5 dev set. Although \method{} has three separate modules that run sequentially, it is $3.5$ times faster than \baselineaif{} at inference time on average. We attribute this to the reduced inference candidate search space due to question augmentation, and smaller input size to the T5 component (refer to Section \ref{sec:method_computation} for details). Please refer to Appendix \ref{app:runtime_analysis} for more details.

\subsection{Error Analysis}
\label{sec:res_error_analysis}
We further analyze the different errors made by \method{} and \baselineaif{} on 50 randomly sampled errors for each model, from the D0-D3 and the subject robustness dev splits. We manually inspect the proof inferences and compare it with the gold proof to classify the failures. The errors are broadly categorized as follows:
\paragraph{Early stop errors:} This is the most frequent error type for both models, accounting for $80 \%$ and $50\%$ errors in \method{} and \baselinea{}, respectively. This occurs when a model incorrectly generates the stop signal and fails to generate all the required inference to prove a statement. We find that our model makes the majority of the mistakes due to early stopping. This can be possibly fixed by improving the rule selector architecture to better model the stop criteria.
\paragraph{Wrong inference:} This is the second error type, where the inferred conclusion is incorrect based on the predicted proof. This accounts for $20 \%$ and $30\%$ errors in \method{} and \baselinea{}, respectively. We observe that our knowledge composer is makes lesser errors on average compared to the \baselinea{} generative model.
\paragraph{Other generation errors:} \baselinea{} makes around $20\%$ errors where the model generated output does not make sense. For example, it can hallucinate facts that are not present in the theory. Such errors are not interpretable and questions the model's inner-working. \method{} shows no such error, since the proofs are always interpretable in our model due to the causal framework.

Overall, we find that the errors made by \method{} are more interpretable than \baselinea{}, since we can pin-point which module is at fault. Whereas, in \baselinea{}, it is sometimes hard to understand the source of errors. This feature also makes our framework easier to debug to potentially fix some components with techniques like data augmentation. Please refer to Appendix \ref{app:error_analysis} for more discussion and examples of errors.

\begin{table}[t]
	\centering
	\resizebox{0.9\columnwidth}{!}{%
		\begin{tabular}{p{7cm}p{2cm}}
			\toprule
			\textbf{Input} & \textbf{Output} \\
			\midrule
			$s_1$: If someone is blue then they are quiet. $s_2$: Chris is blue. & Chris is quiet. ($s_1$, $s_2$) \\
			\midrule
			$s_1$: If someone is blue then they are quiet. $s_2$: Chris is blue. {\color{blue} $s_3$: Steve is blue.} & {\color{red} Dave} is quiet. ($s_1$, $s_2$) \\
			\midrule
			$s_1$: If someone is blue then they are quiet. {\color{blue} $s_2$: Quiet people are cold.} $s_3$: Chris is blue. $s_4$: Steve is blue. {\color{blue} $s_5$: Chris is white.} & {\color{red} Dave} is quiet. ($s_1$, $s_4$) \\
			\bottomrule
		\end{tabular}%
	}
	\caption{\label{tab:results_input_ablation_proofwriter} Examples of inferences made by \baselinea{}. {\color{blue} Blue} text denotes incrementally added sentences in the theory and {\color{red} red} text denotes an error. The $(\cdot)$ is the generated proof. Refer to Section \ref{sec:res_input_ablation} for more details.}
\end{table}

\subsection{\baselinea{} Input Ablation}
\label{sec:res_input_ablation}
A key goal of \method{} is to \textit{explicitly} ensure causality from the rule/facts selection step (proof generation) to the reasoning step (intermediate inference generation). This is essential for a reasoning method using forward chaining to solve a deductive reasoning task \footnote{Forward chaining is described as repeated application of \textit{modus ponens} \cite{forward_vs_backward_chaining}, which requires at least two premises to then logically conclude an inference.}. To understand if \baselinea{}, which uses forward chaining, implicitly does this ``select-then-reason'' within the model, we perform the following case study: We sample theories from our subject perturbation dataset where \baselinea{} made errors, and manually evaluate the model on inputs with all irrelevant rules/facts deleted. Next we sequentially start adding back the deleted rules/facts to see if the output still remains valid. As shown in Table \ref{tab:results_input_ablation_proofwriter}, we see that \baselinea{} generates a correct inference for the first row which uses just the essential part of the theory required to generate the conclusion, and starts making errors as more sentences are included. Some more examples are shown in Table \ref{tab:results_proofwriter_extra_theory} in Appendix. This shows that internally \baselinea{} is unable to faithfully perform the ``select-then-reason'' steps for larger theories. In contrast, \method{} explicitly separates these steps, leading to a faithful reasoning model.
\section{Related Works}

\paragraph{Reasoning in Text}
Reasoning in text is a well studied problem in NLP. Natural Language Inference (NLI) \cite{nli_dataset} is one of the most prominent tasks that require reasoning over text to answer if a statement is entailed, contradicted, or neutral, given a hypothesis. More recently, datasets like HotpotQA \cite{yang2018hotpotqa}, bAbI \cite{babi_dataset}, QuaRTz \cite{tafjord-etal-2019-quartz}, ROPES \cite{lin-etal-2019-reasoning}, CLUTRR \cite{sinha-etal-2019-clutrr}, etc., have studied different aspects of reasoning over textual inputs. These tasks usually require implicit reasoning, where the model needs to internally infer the rules required to solve the task. In contrast, RuleTaker \cite{ruletaker} deals with explicit reasoning (also known as deductive reasoning).

\paragraph{Proof Generation}
Recently, some works have been addressing the problem of proof generation from an NL-based theory. Prover \cite{prover} trains a RoBERTa-based model that predicts nodes and edges of the proof graph. ProofWriter \cite{proofwriter} is a T5-based \cite{raffel2019exploring} model, that iteratively generates one-hop conclusions and proofs from a theory. Another work MultiProver \cite{multiprover}, generates multiple possible proofs for a statement. While we study the same problem of proof generation similar to these works, we develop a more faithful and robust model designing a modular system for proof generation.

\paragraph{Formal Reasoning}
There are some prior works that try to solve the problem of entailment prediction by first parsing the formal language from text. Neural Theorem Prover \cite{ntp_paper,nlprolog} uses neural networks to parse the formal logic from natural language and then reason over them. While this approach is more symbolic, it can lead to many challenges while parsing  \cite{semantic_parsing_survey}. The proof generation setting considered here bypasses this step and directly reasons over the given natural language text making it more useful in downstream applications.

\paragraph{Model Interpretability}
With the advent of pre-trained language models (BERT \cite{devlin2018bert}, RoBERTa \cite{liu2019roberta}, etc.), there has been an increasing trend on solving various reasoning tasks with high accuracy. Faithfulness of such models \cite{jacovi-goldberg-2020-towards} aims to understand whether the models are actually learning to solve the task or rather depending on some shortcut patterns. Saliency-based explanations \cite{sundararajan2017axiomatic,shap,murdoch2018beyond,sanyal-ren-2021-discretized} mainly focus on identifying the important phrases in the input text that helped the model in solving a task. In contrast, the task of proof generation focuses on generating a deductive chain of reasoning from the given theory to the concluded statement. Thus, proof chains are easier to understand for end users, making it more useful to debug any systematic model errors.

\paragraph{Causal Reasoning}
The study of causality and causal reasoning models \cite{causal_pearl_scm,causal_pearl_graphical,causal_bernhard_review} has been prevalent in machine learning. It has been applied in various domains such as algorithmic fairness \cite{causal_algo_fairness}, gender bias mitigation \cite{causal_gender_bias}, robustness from spurious correlations \cite{causal_peter_robustness,causal_veitch_robustness}, counterfactual explanations \cite{feder2021causalm}, etc. Causality in NLP is particularly important to learn models that go beyond exploiting correlations and to improve their overall faithfulness \cite{causal_nlp_review}.
\section{Conclusion}
\vspace{-0.1cm}
In this paper, we proposed \method{}, a faithful and robust deductive reasoning model based on three modular components: rule selection, fact selection, and knowledge composition. \method{} ensures causality from proof generation to entailment prediction by design. We established the effectiveness of our approach through experiments on testing robustness to language variations and demonstrating the interpretability of the errors made by our model. We also show that \method{} is faster and more precise at deductive reasoning than prior baselines.

\section*{Acknowledgments}
This research is supported in part by the Office of the Director of National Intelligence (ODNI), Intelligence Advanced Research Projects Activity (IARPA), via Contract No. 2019-19051600007, the DARPA MCS program under Contract No. N660011924033, the Defense Advanced Research Projects Agency with award W911NF-19-20271, NSF IIS 2048211, NSF SMA 1829268, and gift awards from Google, Amazon, JP Morgan and Sony. We would like to thank all the collaborators in USC INK research lab for their constructive feedback on the work.

\bibliography{custom}
\bibliographystyle{acl_natbib}

\clearpage

\appendix
\section{Depth Dataset Details}
\label{app:depth_dataset}
For training and evaluation of \method{} and \baselinea{}, we use the D* datasets and the ParaRules dataset \cite{ruletaker}. The statistics of these datasets are shown in Table \ref{tab:dataset_stats_pw} which includes the number of theories, the total number of questions across all theories, and the number of conclusions per theory. The statistics are broken down split wise. We use the same splits of train/dev/test as provided in the original datasets \cite{ruletaker,proofwriter}. All the dataset sources are properly cited and used according to the release license.

\begin{table}[t]
    \centering
	\resizebox{\columnwidth}{!}{%
    \begin{tabular}{p{1.2cm}p{0.8cm}P{1.3cm}P{1.3cm}P{2.5cm}}
    \toprule
        \textbf{Dataset} & \textbf{Split} & \textbf{Number of Theories} & \textbf{Number of Questions} & \textbf{Number of Conclusions per Theory (min/mean/max)} \\
        
        \midrule
         & train & 18889 & 69906 & 0/0.81/18 \\
        D0 & dev & 2700 & 10070 & 0/0.81/14 \\
         & test & 5389 & 20024 & 0/0.8/12 \\

		\midrule
         & train & 9008 & 69616 & 1/1.69/13 \\
        D1 & dev & 1318 & 10188 & 1/1.7/14 \\
         & test & 2607 & 20210 & 1/1.7/12 \\

		\midrule
         & train & 6330 & 70076 & 2/3.15/14 \\
        D2 & dev & 909 & 10094 & 2/3.09/12 \\
         & test & 1794 & 19840 & 2/3.11/14 \\

		\midrule
         & train & 4816 & 69388 & 3/4.81/16 \\
        D3 & dev & 719 & 10302 & 3/4.73/14 \\
         & test & 1405 & 20346 & 3/4.72/15 \\

		\midrule
         & train & 3322 & 69810 & 5/9.12/21 \\
        D5 & dev & 482 & 10190 & 5/9.13/21 \\
         & test & 948 & 20030 & 5/9.08/21 \\

		\midrule
         & train & 1681 & 28010 & 3/4.25/14 \\
        Pararules & dev & 240 & 4004 & 3/4.53/13 \\
         & test & 482 & 8008 & 3/4.24/11 \\
        
        \bottomrule
    \end{tabular}
    }
	\caption{\label{tab:dataset_stats_pw} Statistics of datasets introduced by \citealp{proofwriter} with the number of theories, questions and conclusions per theory for all three splits of each dataset. The splits are kept the same as the original dataset. Please refer to Appendix \ref{app:depth_dataset} for more details.
	}
\end{table}

\begin{table}[t]
    \centering
	\resizebox{\columnwidth}{!}{%
    \begin{tabular}{p{2.5cm}p{0.8cm}P{1.3cm}P{1.3cm}P{2.5cm}}
    \toprule
        \textbf{Dataset} & \textbf{Split} & \textbf{Number of Theories} & \textbf{Number of Questions} & \textbf{Number of Conclusions per Theory (min/mean/max)} \\

		\midrule
        & train & 28896 & 416328 & 3/4.81/16 \\
        Subject & dev & 4314 & 61812 & 3/4.73/14 \\
        ~ & test & 8430 & 122076 & 3/4.72/15 \\

		\midrule
        & train & 28866 & 415848 & 3/4.81/16 \\
        Attribute & dev & 4314 & 61812 & 3/4.73/14 \\
        ~ & test & 8415 & 121836 & 3/4.73/15 \\

		\midrule
        \multirow{3}{*}{Subject+Attribute} & train & 28866 & 415848 & 3/4.81/16 \\
         & dev & 4314 & 61812 & 3/4.73/14 \\
         & test & 8415 & 121836 & 3/4.73/15 \\
        \bottomrule
    \end{tabular}
    }
	\caption{\label{tab:dataset_stats_robustness} Statistics of datasets introduced in this paper with the number of theories, questions and conclusions per theory for all three splits of each dataset. We use these datasets to quantify the robustness of \methodsmall{} and compare it with baselines. Please refer to Appendix \ref{app:robustness_dataset} for more details.}
\end{table}

\section{Robustness Dataset Details}

\label{app:robustness_dataset}
The robustness dataset is created by replacing all subjects (attributes, subject+attributes) in the D3 dataset with unseen subjects (attributes, subject+attributes) to create the subject (attribute, subject+attributes) robustness set. For this, we first curate new sets of subjects and attributes to be used as a global pool to sample from while replacing existing subjects and attributes from the theory. These sets are detailed below:
\paragraph{Subject proper name pool:} \{`George', `Paul', `Ronald', `Emma', `Magnus', `Timothy', `Chris', `Molly', `Diana', `Joseph', `Becky', `Kurt', `Ivan', `Steve', `Laura', `Oliver', `Adam', `Larry'\}
\paragraph{Subject common name pool:} \{`mother', `father', `baby', `child', `toddler', `teenager', `grandmother', `student', `teacher', `alligator', `cricket', `bird', `wolf', `giraffe', `dinosaur', `thief', `soldier', `officer', `artist', `shopkeeper', `caretaker', `janitor', `minister', `salesman', `saleswoman', `runner', `racer', `painter', `dresser', `shoplifter'\}
\paragraph{Attribute pool:} \{`maroon', `brown', `black', `orange', `cordial', `friendly', `adorable', `old', `soft', `violent', `intelligent', `square', `warm', `large', `cylindrical', `spherical', `tiny', `microscopic', `brilliant', `noisy', `playful', `tender', `gracious', `patient', `funny', `hilarious', `thorny', `sensitive', `diplomatic', `thoughtful'\}

Then, for each theory in the D3 dataset, we replace \textit{all} the subjects in the theory with randomly sampled subjects (without replacement) from the candidate set to create a perturbed theory. We perform this replacement operation to generate five different perturbed theories. These perturbed theories are called equivalence set. Note that the only change in each theory in an equivalence set is the subjects being replaced by some randomly sampled subjects. For example, ``cat'' in the original theory might be replaced by ``child'' in one perturbation, and with ``teacher'' in yet another perturbation. We follow the same procedure to create attribute and subject+attribute robustness sets.

The statistics for these robustness datasets are shown in Table \ref{tab:dataset_stats_robustness} which includes the dataset name depicting the perturbation type (subject, attribute or subject+attribute), number of theories, the total number of questions across all theories, and the number of conclusions per theory. Please note that one theory has multiple questions in general, and it is possible to have conclusions that are not a part of these questions, but can be deduced from the given theory. Each split of the original dataset is perturbed separately as described above, to create the new datasets.

\begin{table}[t]
	\centering
	\resizebox{\columnwidth}{!}{%
		\begin{tabular}{ccccccc}
			\toprule
			& \multicolumn{3}{c}{\textbf{Entailment Accuracy}}	& \multicolumn{3}{c}{\textbf{Proof Accuracy}}	\\
			\cmidrule(r){2-4} \cmidrule(r){5-7}
			$d$ & \baselineaa{} & \baselineai{} & \methodsmall{} & \baselineaa{} & \baselineai{} & \methodsmall{}	\\
			\midrule
			N/A & 97.4 & 99.2 & 99.4 & 97.4 & 99.2 & 99.4 \\
			0 	& 100.0 & 100.0 & 100.0 & 100.0 & 100.0 & 100.0 \\
			1 	& 99.9 & 99.1 & 99.5 & 99.3 & 97.5 & 99.2 \\
			2 	& 99.7 & 98.9 & 98.5 & 97.6 & 96.4 & 96.1 \\
			3 	& 99.7 & 98.4 & 93.4 & 91.2 & 95.5 & 85.5 \\
			4 	& 99.5 & 97.5 & 88.8 & 46.9 & 93.4 & 77.4 \\
			5 	& 98.9 & 96.5 & 79.2 & 24.4 & 82.3 & 68.1 \\
			\midrule
			All & 98.7 & 98.8 & 95.9 & 85.6 & 96.4 & 92.7 \\
			\bottomrule
		\end{tabular}%
	}
	\caption{\label{tab:results_d03_d5} D5 dataset depth-wise performance comparison of \method{} trained on D0-D3 with \baselineaaf{} and \baselineaif{} trained on D3 and D0-D3 respectively. Baseline results are copied from \citet{proofwriter}. Refer to Section \ref{sec:res_depth} and Appendix \ref{app:depth_generalization} for more details. 
	}
\end{table}

\section{Generalization to Reasoning Depths}
\label{app:depth_generalization}
In this section, we experiment with a setting where models are trained on depths less than or equal to 3 (i.e., $d \leq 3$) and tested on D5 dataset that contains statements that require reasoning up to depth 5 (i.e., $d \leq 5$). Here, we test the generalization of the models to reasoning depths that are \textit{unseen} at training time. These results are shown in Table \ref{tab:results_d03_d5}. From this table, we observe that overall our model performs significantly better than \baselineaaf{} on proof accuracy ($+7.5\%$), but has a lower performance compared to \baselineaif{} ($-3\%$). This shows that compared to \baselineaif{}, our models are weaker at generalizing to unseen reasoning depths. This happens majorly because our rule selector tends to stop the inference iterations earlier, which means some essential inferences are not generated by the model. Thus, this leads to lower performance with increasing reasoning depths.

But, we make another interesting observation here. The drops in entailment and proof accuracy with increasing depths are similar for \method{}. For instance, considering the performance drops between $d=4$ to $d=5$, \method{} has $\sim 9.5 \%$ drop in both entailment and proof accuracy. In contrast, \baselineaaf{} and \baselineaif{} drops approximately $22\%$ and $11 \%$, respectively in proof accuracy for a mere $1 \%$ drop in entailment accuracy. This raises some concern on the causality of the proof generation process used for entailment prediction in these models, since it seems like the answer prediction and proof generation are not dependent via the same reasoning paths. In contrast, our causal framework grounds the entailment prediction to the proofs and this leads to more consistent performance variations in \method{}.

\begin{table}[t]
	\centering
	\resizebox{\columnwidth}{!}{%
		\begin{tabular}{lcccc}
			\toprule
			& \multicolumn{2}{c}{\textbf{Entailment Accuracy}}	& \multicolumn{2}{c}{\textbf{Proof Accuracy}}	\\
			\cmidrule(r){2-3} \cmidrule(r){4-5}
			$d$ & \baselineai{} & \methodsmall{} & \baselineai{} & \methodsmall{}	\\
			\midrule
			N/A & 98.9 & 99.3 & 98.9 & 99.3 \\
			0 & 99.9 & 100.0 & 99.9 & 100.0 \\
			1 & 79.1 & 96.0 & 78.8 & 95.7 \\
			2 & 76.6 & 93.4 & 73.4 & 91.4 \\
			3 & 72.7 & 89.8 & 67.8 & 85.7 \\
			\midrule
			All & 89.6 & 96.8 & 88.4 & 95.9 \\
			\bottomrule
		\end{tabular}%
	}
	\caption{\label{tab:results_d03_subject_depth} Comparison of \method{} with \baselineaif{} trained on D0-D3 and tested on subject robustness dataset. Baseline results are generated using the checkpoint provided by the authors. For more details, please refer to Appendix \ref{app:robustness_depth}.}
\end{table}

\begin{table}[t]
	\centering
	\resizebox{\columnwidth}{!}{%
		\begin{tabular}{lcccc}
			\toprule
			& \multicolumn{2}{c}{\textbf{Entailment Accuracy}}	& \multicolumn{2}{c}{\textbf{Proof Accuracy}}	\\
			\cmidrule(r){2-3} \cmidrule(r){4-5}
			$d$ & \baselineai{} & \methodsmall{} & \baselineai{} & \methodsmall{}	\\
			\midrule

			N/A & 99.6 & 99.1 & 99.6 & 99.1 \\
			0 & 100.0 & 100.0 & 100.0 & 100.0 \\
			1 & 96.4 & 96.0 & 96.2 & 95.6 \\
			2 & 95.1 & 93.7 & 94.1 & 91.3 \\
			3 & 93.9 & 89.5 & 92.1 & 84.1 \\
			\midrule
			All & 97.8 & 96.7 & 97.4 & 95.6 \\

			\bottomrule
		\end{tabular}%
	}
	\caption{\label{tab:results_d03_attribute_depth} Comparison of \method{} with \baselineaif{} trained on D0-D3 and tested on attribute robustness dataset. Baseline results are generated using the checkpoint provided by the authors. For more details, please refer to Appendix \ref{app:robustness_depth}.}
\end{table}

\section{Robustness to Perturbed Theories}
\label{app:robustness_depth}
Here, we show the detailed depth-wise performance of \method{} and \baselineaif{} trained on D0-D3 dataset and evaluated on different robustness datasets as described in Section \ref{sec:setup}. The results for subject, attribute, and subject+attribute robustness evaluations are shown in Tables \ref{tab:results_d03_subject_depth}, \ref{tab:results_d03_attribute_depth}, and \ref{tab:results_d03_sub_attr_depth}, respectively. We observe that \baselineaif{} performs significantly worse compared to \method{} on subject robustness. The results on subject+attribute robustness are mostly comparable, while in attribute robustness our model performs worse. The drop in performance show that both the models are sensitive to attributes in the theory to varying degree. But the strong sensitivity of \baselineaif{} to the subject perturbations is questionable, since the causality of the model's reasoning process seems to be compromised because the model learns some spurious correlations using the subjects.

In another setting, we train different components of our model on the robustness data and check if that leads to some performance gains. These results are reported in Table \ref{tab:results_train_robust}. We find that it is indeed possible to improve the performance of individual components of our model by robust data augmentation. This also indicates that our individual components are flexible to intervention by data augmentation. Such abilities are lacking in \baselinea{}.

\begin{table}[t]
	\centering
	\resizebox{\columnwidth}{!}{%
		\begin{tabular}{lcccc}
			\toprule
			& \multicolumn{2}{c}{\textbf{Entailment Accuracy}}	& \multicolumn{2}{c}{\textbf{Proof Accuracy}}	\\
			\cmidrule(r){2-3} \cmidrule(r){4-5}
			$d$ & \baselineai{} & \methodsmall{} & \baselineai{} & \methodsmall{}	\\
			\midrule

			N/A & 98.6 & 98.9 & 98.6 & 98.9 \\
			0 & 100.0 & 100.0 & 100.0 & 100.0 \\
			1 & 91.3 & 94.0 & 90.9 & 93.6 \\
			2 & 89.2 & 90.3 & 85.9 & 87.8 \\
			3 & 85.9 & 85.9 & 80.0 & 80.5 \\
			\midrule
			All & 94.8 & 95.4 & 93.4 & 94.3 \\
			\bottomrule

		\end{tabular}%
	}
	\caption{\label{tab:results_d03_sub_attr_depth} Comparison of \method{} with \baselineaif{} trained on D0-D3 and tested on subject+attribute robustness dataset. Baseline results are generated using the checkpoint provided by the authors. For more details, please refer to Appendix \ref{app:robustness_depth}.}
\end{table}

\begin{table}[t]
	\centering
	\resizebox{\columnwidth}{!}{%
		\begin{tabular}{lcccc}
			\toprule
			\multirow{2}{*}{\textbf{Trained module}} & \multicolumn{2}{c}{\textbf{Subj Perturbation}} & \multicolumn{2}{c}{\textbf{Attr Perturbation}} \\
			\cmidrule(r){2-3} \cmidrule(r){4-5}
			& EA & PA	& EA & PA \\
			\midrule
			Base (trained on D0-D3) & 96.8 & 95.9 & 96.7 & 95.6 \\
			\midrule
			Rule selector (RS) & 97.3 & 96.7 & 96.1 & 95.3 \\
			Fact selector (FS) & 96.7 & 95.8 & 96.6 & 94.6 \\
			Knowledge composer (KC) & 98.5 & 97.6 & 98.2 & 97.1 \\
			RS + FS + KC & 99.1  & 98.5 & 98.7 & 97.2 \\

			\bottomrule
		\end{tabular}%
	}
	\caption{\label{tab:results_train_robust} Comparison of variants of \method{} where different components (RS, FS, KC) and their combinations are trained and tested on subject robustness datasets. EA and PA refers to entailment accuracy and proof accuracy respectively. Please refer to Appendix \ref{app:robustness_depth} for more details.}
\end{table}

\begin{figure}[t]
\vspace{-0.3cm}
	\centering
	\includegraphics[width=0.85\columnwidth]{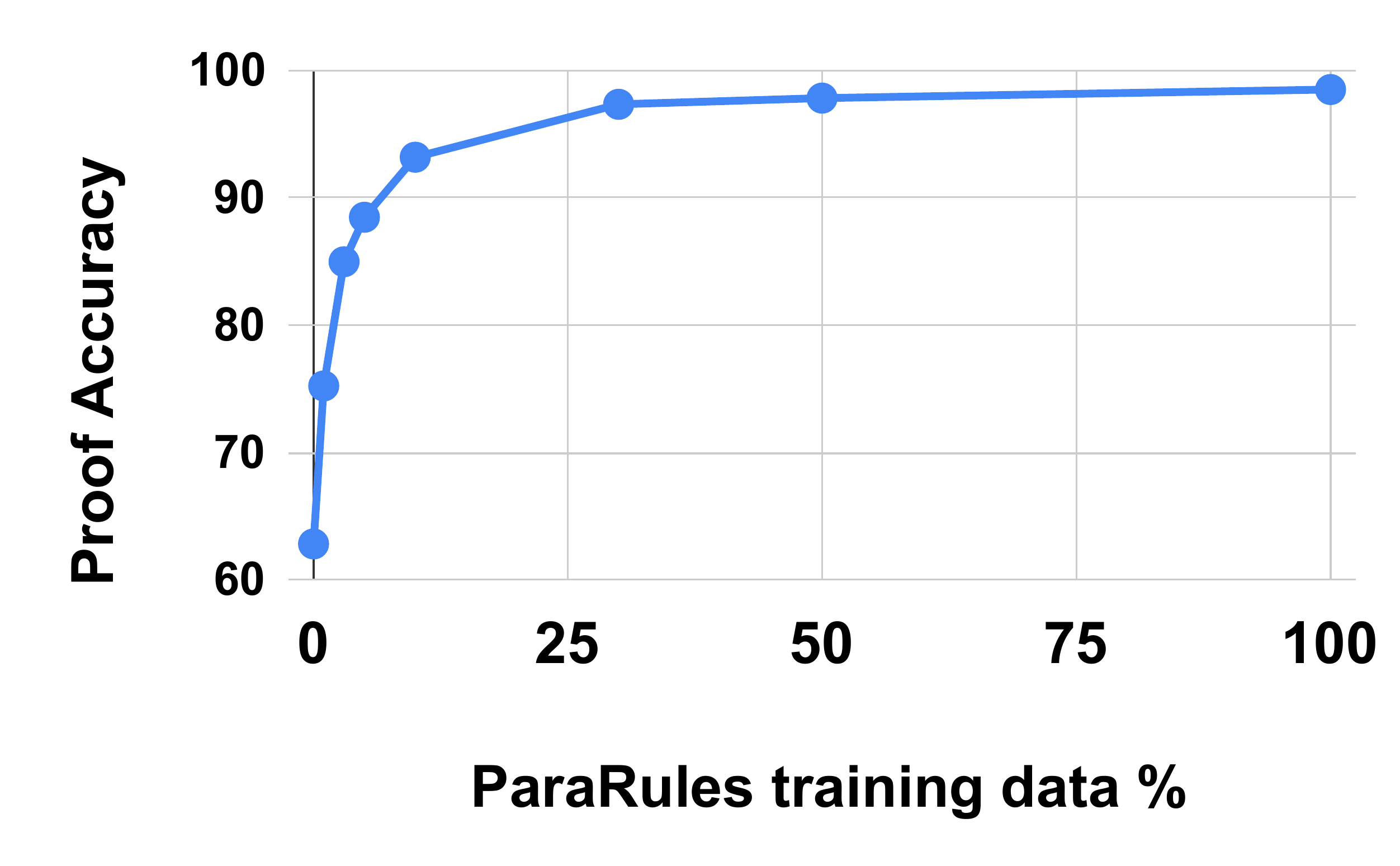}
	\vspace{-0.3cm}
	\caption{\small \label{fig:para_low_resource}Proof Accuracy of \method{} when tested on ParaRules while using limited amount of ParaRules along with D0-D3 for training. See Appendix ~\ref{app:res_pararules} for more details.}
	\vspace{-0.3cm}
\end{figure}

\section{Generalization to paraphrased theories}
\label{app:res_pararules}
Here we test the ability of our model to generalize to unseen language in ParaRules by using limited training supervision. To test this, we first train our model on D0-D3 dataset and test it on the ParaRules dataset. This is a zero-shot evaluation on an unseen language form. In Figure \ref{fig:para_low_resource} we observe that the performance is significantly worse on this setting as expected. We also evaluated a checkpoint of \baselineaif{} trained on D0-D3 which achieves a similar performance of $62.13\%$ entailment accuracy \footnote{data-augmented training results for \baselinea{} are not reported in the figure since the training code is not available}. Next, we gradually start adding portions of ParaRules, along with the D0-D3 data, to the training dataset. We find that \method{} can quickly achieve reasonable performance using even $10\%$ additional data. This shows that our modularized approach is also efficient in adapting to unseen theories with limited data supervision. For more comparisons with models trained on ParaRules, please refer to Appendix \ref{app:pararules}.

\begin{table}[t]
	\centering
	\resizebox{\columnwidth}{!}{%
		\begin{tabular}{lcccc}
			\toprule
			& \multicolumn{2}{c}{\textbf{Entailment Accuracy}}	& \multicolumn{2}{c}{\textbf{Proof Accuracy}}	\\
			\cmidrule(r){2-3} \cmidrule(r){4-5}
			$d$ & \baselineaa{} [T5-11B] & \methodsmall{} & \baselineaa{} [T5-11B] & \methodsmall{}	\\
			\midrule
			0 & 99.9 & 100.0 & 99.9 & 100.0 \\
			1 & 99.3 & 99.6 & 99.3 & 99.6 \\
			2 & 98.3 & 97.6 & 97.7 & 97.4 \\
			3 & 98.2 & 95.4 & 96.5 & 95.1 \\
			4 & 91.5 & 91.6 & 83.1 & 91.6 \\
			\midrule
			All & 99.1 & 98.7 & 98.5 & 98.6 \\
			\bottomrule
		\end{tabular}%
	}
	\caption{\label{tab:results_para} Comparison of \method{} with \baselineaaf{} [T5-11B] when trained on D3+ParaRules and tested on ParaRules. Results for \baselineaaf{} [T5-11B] are copied from the paper. Please refer to Appendix \ref{app:pararules} for more details.}
\end{table}

\section{Results on ParaRules training}
\label{app:pararules}
Following \cite{proofwriter}, we compare the performance of \baselineaaf{} and \method{} on the ParaRules dataset, when trained on a combined partition of D3 and ParaRules train set. The ParaRules dataset contains complex linguistic expressions in the theories that are more realistic than D* dataset theories, making it a more challenging dataset. These results are shown in Table \ref{tab:results_para}, with a reasoning depth breakdown as before. We note that numbers for \baselineaif{} are not reported in the paper, and no trained checkpoint is available either, so we omit it from our comparisons. Also, the reported results for \baselineaaf{} are from evaluating a T5-11B model while ours is a T5-large model. Here, we see that our model performs better at higher depths compared to the baseline which demonstrates that \method{} is better at handling paraphrases.

\begin{figure}[t]
	\centering
	\begin{subfigure}{.5\textwidth}
		\centering
		\includegraphics[width=0.9\columnwidth]{figures/budget_dep_1.pdf}
		\caption{Depth 1}
	\end{subfigure}%
	\\
	\begin{subfigure}{.5\textwidth}
		\centering
		\includegraphics[width=0.9\columnwidth]{figures/budget_dep_3.pdf}
		\caption{Depth 3}
	\end{subfigure}%
	\\
	\begin{subfigure}{.5\textwidth}
		\centering
		\includegraphics[width=0.9\columnwidth]{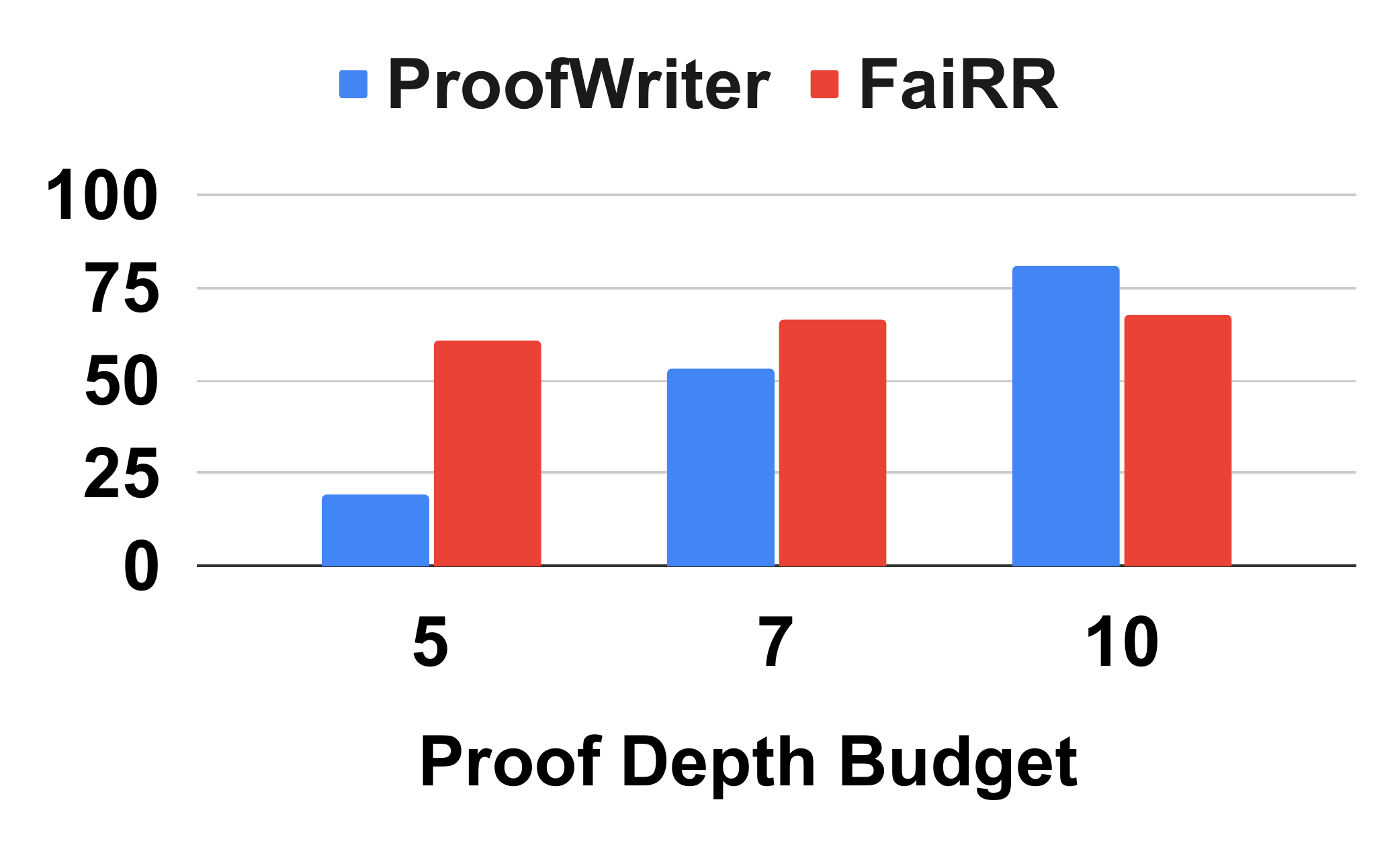}
		\caption{Depth 5}
	\end{subfigure}%
	\caption{\label{fig:budget} Depth-wise comparison of \baselineaif{} and \method{} (both trained on D0-3 dataset) on limited inference budgets. Please refer to Appendix \ref{app:budget_analysis} for details.
	}
\end{figure}

\section{Inference Budget Analysis}
\label{app:budget_analysis}
In the inference budget analysis, we compare the performance of \method{} and \baselinea{} under an inference budget constraint, i.e., we restrict the total number of intermediate conclusions that can be produced by both models. We perform this analysis on three different depth datasets ($d = \{1, 3, 5\}$) and upper bound the number of inferences by $B = \{1, 3, 5, 7, 10\}$. We ensure that the budget is at least equal to the depth of the statements under consideration since proving a statement requires a model to generate inferences equal to at least the depth. From Figure \ref{fig:budget} we observe that for all depths \method{} consistently outperforms \baselinea{} on lower budgets. Only when the budget increases to 10, \baselinea{} compares with or sometimes outperforms our model. This analysis demonstrates that \method{} performs a prioritized generation of conclusions that are relevant to the statement, which can be useful in scenarios with limited inference budgets.

\begin{table}[t]
	\centering
	\resizebox{0.7\columnwidth}{!}{%
		\begin{tabular}{lcc}
			\toprule
			$d$ & \baselineaif{} & \method{} \\
			\midrule
			0 & 1.01 & 0.12 \\
			1 & 1.01 & 0.20 \\
			2 & 1.00 & 0.28 \\
			3 & 1.01 & 0.36 \\
			4 & 1.01 & 0.46 \\
			\midrule
			Avg & 1.01 & 0.28 \\
			\bottomrule
		\end{tabular}%
	}
	\caption{\label{tab:results_runtime} Evaluation runtime (in hours) of \method{} and \baselineaif{}. Please refer to Appendix \ref{app:runtime_analysis} for more details.}
\end{table}

\section{Runtime Analysis}
\label{app:runtime_analysis}
For inference runtime analysis, we time the evaluation of both \method{} and \baselineaif{} on D5 dev set. Note that D5 dataset contains statements that require \textit{at most} five reasoning steps to generate an answer. The runtime for both methods are shown in Table \ref{tab:results_runtime}. These results were obtained by running the inference algorithm on NVIDIA GeForce RTX 2080 Ti GPUs for both models. We observe that \baselineaif{} has an almost constant runtime since it always generates all possible inferences for a theory. In contrast, our runtime increases almost linearly with increasing depth. On average, \method{} is $3.5$ times faster at inference than \baselineaif{}.

\begin{table*}[t]
	\centering
	\resizebox{0.9\textwidth}{!}{%
		\begin{tabular}{p{3cm}p{10cm}p{5cm}}
			\toprule
			\textbf{Error Type} &\textbf{Input ( \textit{format: complete theory})} & \textbf{Output ( \textit{format: \$answer\$ = conclusion ; \$proof\$ = \# rule \& facts})} \\
			\midrule
			Wrong inference & \$answer\$ ; \$proof\$ ; \$question\$ = What is one single-hop inference? ; \$context\$ = sent1: Cold people are red. sent2: If someone is quiet then they are cold. sent3: Smart, blue people are quiet. sent4: If Paul is blue then Paul is furry. \textbf{sent5}: If someone is blue then they are quiet. sent6: If Steve is furry and Steve is smart then Steve is white. sent7: If Chris is red and Chris is furry then Chris is blue. sent8: Quiet people are cold. \textbf{sent9}: Chris is blue. sent10: Chris is white. sent11: Ronald is red. sent12: Paul is quiet. sent13: Paul is smart. sent14: Steve is blue. sent15: Steve is furry. sent16: Steve is red. sent17: Steve is smart. sent18: Steve is white. & \$answer\$ = {\color{red} Dave is quiet.} ; \$proof\$ = \# sent5 sent9 \\
			\midrule
			Wrong inference & \$answer\$ ; \$proof\$ ; \$question\$ = What is one single-hop inference? ; \$context\$ = sent1: Cold people are red. \textbf{sent2}: If someone is quiet then they are cold. sent3: Smart, blue people are quiet. sent4: If Paul is blue then Paul is furry. sent5: If someone is blue then they are quiet. sent6: If Steve is furry and Steve is smart then Steve is white. sent7: If Chris is red and Chris is furry then Chris is blue. sent8: Quiet people are cold. sent9: Chris is blue. sent10: Chris is white. sent11: Ronald is red. sent12: Paul is quiet. sent13: Paul is smart. sent14: Steve is blue. sent15: Steve is furry. sent16: Steve is red. sent17: Steve is smart. sent18: Steve is white. \textbf{sent19}: Dave is quiet. & \$answer\$ = Dave is {\color{red} not a cat}. ; \$proof\$ = \# sent2 sent19 \\
			\midrule
			Generation error - output format violated & \$answer\$ ; \$proof\$ ; \$question\$ = What is one single-hop inference? ; \$context\$ = sent1: If someone likes the grandmother then the grandmother eats the artist. sent2: If someone eats the artist then the artist is young. sent3: If the artist eats the grandmother then the artist is cold. sent4: If someone eats the artist and the artist eats the grandmother then the grandmother eats the artist. sent5: If someone is red and big then they like the grandmother. \textbf{sent6}: If someone visits the grandmother then they are red. sent7: If someone is rough then they eat the grandmother. sent8: If someone likes the artist then they visit the grandmother. sent9: The artist likes the grandmother. sent10: The grandmother eats the artist. sent11: The grandmother is big. sent12: The grandmother is cold. sent13: The grandmother is young. sent14: The grandmother likes the artist. \textbf{sent15}: The grandmother visits the artist. sent16: The artist is young. & \$answer\$ = {\color{red} The red sent6 sent15} \\
			\midrule
			Generation error - hallucinating facts & \$answer\$ ; \$proof\$ ; \$question\$ = What is one single-hop inference? ; \$context\$ = sent1: All young things are smart. \textbf{sent2}: All smart, young things are nice. sent3: If something is young and nice then it is quiet. sent4: Becky is big. sent5: Becky is green. sent6: Becky is young. sent7: Larry is big. sent8: Larry is quiet. sent9: Larry is red. sent10: Larry is young. \textbf{sent11}: Bob is smart. & \$answer\$ = Bob is nice. ; \$proof\$ = \# sent2 \& sent11 {\color{red} sent16} \\
			\midrule
			Generation error - repeated facts & \$answer\$ ; \$proof\$ ; \$question\$ = What is one single-hop inference? ; \$context\$ = sent1: If Adam is green then Adam is big. sent2: Green things are rough. sent3: If Joseph is cold then Joseph is big. sent4: All blue things are cold. sent5: All green things are rough. sent6: All big things are kind. sent7: If something is rough and big then it is green. sent8: All kind, blue things are cold. sent9: Joseph is blue. sent10: Adam is blue. sent11: Molly is cold. sent12: Joseph is cold. \textbf{sent13}: Charlie is cold. & \$answer\$ = Charlie is {\color{red} sent13 sent13} \\
			\midrule
			Generation error - output format violated & \$answer\$ ; \$proof\$ ; \$question\$ = What is one single-hop inference? ; \$context\$ = sent1: Nice things are green. sent2: If Ivan is smart then Ivan is blue. sent3: Green things are smart. sent4: All cold things are nice. sent5: Big things are blue. sent6: If Ronald is nice then Ronald is big. sent7: If something is green and big then it is not kind. \textbf{sent8}: If something is smart and nice then it is kind. \textbf{sent9}: Ronald is smart. sent10: Larry is cold. sent11: Laura is nice. sent12: Ivan is kind. sent13: Larry is nice. sent14: Larry is green. sent15: Bob is smart. & \$answer\$ = {\color{red} Bob} is kind. ; \$proof\$ = \# sent8 \& {\color{red} sent9 sentnice} \\
			\bottomrule
		\end{tabular}%
	}
	\caption{\label{tab:results_error_proofwriter} Examples of inference errors made by \baselineaif{}. Please refer to Appendix \ref{app:error_analysis} for more details.}
\end{table*}

\begin{table*}[t]
	\centering
	\resizebox{0.9\textwidth}{!}{%
		\begin{tabular}{p{3cm}p{10cm}p{3cm}}
			\toprule
			\textbf{Error Type} & \textbf{Input ( \textit{format: [facts], rule})} & \textbf{Inference} \\
			\midrule
			Wrong inference & [the racer needs the janitor.], if someone needs the janitor then the  janitor likes the racer. & the janitor likes the {\color{red} race}. \\
			\midrule
			Wrong inference & [oliver is big.], big people are young. & {\color{red} the} oliver is young. \\
			\midrule
			Wrong inference & [the shoplifter needs the shoplifter., the shoplifter needs the dinosaur.], if something needs the dinosaur and it needs the shoplifter then it is round. & the {\color{red} shop} is round. \\
			\bottomrule
		\end{tabular}%
	}
	\caption{\label{tab:results_error_causalreasoner} Examples of inference errors made by \method{} while composing rules and facts using the knowledge composer. Please refer to Appendix \ref{app:error_analysis} for more details.}
\end{table*}

\begin{table*}[t]
	\centering
	\resizebox{0.9\textwidth}{!}{%
		\begin{tabular}{p{11.5cm}p{5cm}P{2cm}}
			\toprule
			\textbf{Input ( \textit{format: complete theory})} & \textbf{Output ( \textit{format: \$answer\$ = conclusion ; \$proof\$ = \# rule \& facts})} & \textbf{Output \newline \cmark:correct \newline \xmark:incorrect}  \\
		
	    	\midrule
            \textbf{Example 1} & & \\
            \midrule
            
			\$answer\$ ; \$proof\$ ; \$question\$ = What is one single-hop inference? ; \$context\$ = sent1: If someone is blue then they are quiet. sent2: Chris is blue.
			& \$answer\$ = Chris is quiet. ; \$proof\$ = \# sent1 sent2 
			& \cmark \\
			\midrule

			\$answer\$ ; \$proof\$ ; \$question\$ = What is one single-hop inference? ; \$context\$ = sent1: If someone is blue then they are quiet. sent2: Chris is blue. {\color{blue} sent3: Steve is blue.}
			& \$answer\$ = {\color{red} Dave is quiet.} ; \$proof\$ = \# sent1 sent2 
			& \xmark \\
			\midrule

			\$answer\$ ; \$proof\$ ; \$question\$ = What is one single-hop inference? ; \$context\$ = sent1: If someone is blue then they are quiet. {\color{blue} sent2: Quiet people are cold.} sent3: Chris is blue. sent4: Steve is blue. {\color{blue} sent5: Chris is white.}
			& \$answer\$ = {\color{red} Dave is quiet.} ; \$proof\$ = \# sent1 sent4 
			& \xmark \\

            \midrule
            \textbf{Example 2} & & \\
			\midrule

			\$answer\$ ; \$proof\$ ; \$question\$ = What is one single-hop inference? ; \$context\$ = sent1: If someone likes the artist then they visit the grandmother. sent2: The grandmother likes the artist.
            & \$answer\$ = The grandmother visits the grandmother. ; \$proof\$ = \# sent1 sent2 
			& \cmark \\
			\midrule

			\$answer\$ ; \$proof\$ ; \$question\$ = What is one single-hop inference? ; \$context\$ = {\color{blue} sent1:  If someone is red and big then they like the grandmother. sent2: If someone visits the grandmother then they are red.} sent3: If someone likes the artist then they visit the grandmother. {\color{blue} sent4: The artist likes the grandmother. sent5: The grandmother eats the artist. sent6: The grandmother is big. sent7: The grandmother is cold.} sent8: The grandmother likes the artist.
            & \$answer\$ = {\color{red} The artist visits the grandmother.} ; \$proof\$ = \# {\color{red} sent3 sent4} 
			& \xmark \\
			\midrule 

            \$answer\$ ; \$proof\$ ; \$question\$ = What is one single-hop inference? ; \$context\$ = {\color{blue} sent1: If someone likes the grandmother then the grandmother eats the artist. sent2: If someone eats the artist then the artist is young. sent3: If the artist eats the grandmother then the artist is cold. sent4: If someone eats the artist and the artist eats the grandmother then the grandmother eats the artist.} sent5: If someone is red and big then they like the grandmother. sent6: If someone visits the grandmother then they are red. {\color{blue} sent7: If someone is rough then they eat the grandmother.} sent8: If someone likes the artist then they visit the grandmother. sent9: The artist likes the grandmother. sent10: The grandmother eats the artist. sent11: The grandmother is big. sent12: The grandmother is cold. {\color{blue} sent13: The grandmother is young.} sent14: The grandmother likes the artist. {\color{blue} sent15: The grandmother visits the artist. sent16: The artist is young.}
            & \$answer\$ = {\color{red} The red sent6 sent15} 
			& \xmark \\

            \midrule
            \textbf{Example 3} & & \\
            \midrule

			\$answer\$ ; \$proof\$ ; \$question\$ = What is one single-hop inference? ; \$context\$ = sent1: All young things are smart. sent2: All smart, young things are nice. sent3: Bob is smart. sent4: Bob is young
            & \$answer\$ = Bob is nice. ; \$proof\$ = \# sent2 \& sent3 sent4 
			& \cmark \\
            \midrule

			\$answer\$ ; \$proof\$ ; \$question\$ = What is one single-hop inference? ; \$context\$ = sent1: All young things are smart. sent2: All smart, young things are nice. {\color{blue} sent3: If something is young and nice then it is quiet. sent4: Becky is big. sent5: Becky is green. sent6: Becky is young. sent7: Larry is big. sent8: Larry is quiet. sent9: Larry is red. sent10: Larry is young.} sent11: Bob is smart.
            & \$answer\$ = {\color{red} Bob is nice.} ; \$proof\$ = \# sent2 \& sent11 {\color{red} sent16} 
			& \xmark \\

			\bottomrule
		\end{tabular}%
	}
	\caption{\label{tab:results_proofwriter_extra_theory} Some more examples of inference errors made by \baselineaif{}. We see that having extra information in the theory than what is required to prove the conclusion leads to errors (shown in {\color{red} red}). Having limited information in the theory reduces errors. Sentences in {\color{blue} blue} depict the sentences which are added to the theory with respect to the row above. Please refer to Section \ref{sec:res_input_ablation} for more details.}
\end{table*}

\section{Error Analysis}
\label{app:error_analysis}
This is a follow-up of Section \ref{sec:res_error_analysis}, where we delve deeper into the error analysis by discussing different error examples and their potential reasons. First, the stop errors are easy to understand. These are cases where the model just decides to stop instead of generating any further conclusions. For our model, this can happen if the rule selector is under confident while selecting rules and it learns that a safer fallback is to stop generating rules. This aspect can probably be improved by a better modeling of the rule selector. We plan to explore this in future works.

Next we look at some of the wrong inferences generated by both models in Tables \ref{tab:results_error_proofwriter} and \ref{tab:results_error_causalreasoner}. We observe that errors made by \method{} are rather naive with small mistakes in the final conclusion (shown in red in Table \ref{tab:results_error_causalreasoner}). In contrast, \baselinea{} tends to generate an invalid conclusion with no relation to the generated proof (rows 1 and 2 in Table \ref{tab:results_error_proofwriter}). It also makes many non-interpretable generation errors where the model's output format is completely violated or the model seems to hallucinate some facts (rows 3-6 in Table \ref{tab:results_error_proofwriter}). Thus, we observe the benefit of our causal framework as the errors are interpretable and more believable. In contrast, errors made by \baselinea{} clearly show that its inference reasoning process can often not rely on the proof at or, or even the generated proof sometimes doesn't make sense.

\section{Comparison with Baselines}
\label{app:comp_baselines}
In this work we compare \method{} with baselines introduced by \cite{proofwriter}. We omit comparisons with both \baselineb{} \cite{prover} and \baselinebmulti{} \cite{multiprover}, since they were trained on a different dataset that makes a closed-world assumption (CWA), whereas we use datasets that make an open-world assumption (OWA). One essential difference between these two datasets are that OWA allows for predicting the truth values as one of \{True,False,Unknown\} while in CWA, any fact that cannot be deduced from the theory is assumed to be false. As a result, in CWA, there are only two possible truth values \{True,False\} for a given statement. This CWA assumption also leads to a specific constraint in the generated proof graphs, where special $NAF$ nodes need to be considered. Please refer to \citet{prover} for more details on this.
Additionally, \baselinebmulti{} \cite{multiprover} has a goal that is different from ours. Specifically, their focus is on generating multiple possible proofs for a given rulebase, and for this their training examples contain multiple gold proofs per instance. In \method{} and ProofWriter \cite{proofwriter}, only one gold proof needs to be generated, making the comparisons a bit unfair. Following \citet{proofwriter}, we report single run numbers for every experiment.

\section{Hyperparameters}
\label{app:hyperparams}
We use RoBERTa-large models \cite{liu2019roberta} to model the rule selector and fact selector in \method{}.
For selecting the best hyperparameters for both these components, we selected the max training epochs in: $\{10, 15, 20\}$, warmup updates in: $\{0.05, 0.1\}$, weight decay in: $\{0.1, 0.01, 0.001\}$, learning rate in: $\{3e\mhyphen6, 5e\mhyphen6, 1e\mhyphen6\}$, and batch size in $\{16, 32\}$.

We use T5 \cite{raffel2019exploring} (T5-large) to model the knowledge composer in \method{} and train it using the default hyperparameters available in the Hugging Face transformers library \cite{wolf-etal-2020-transformers}. All models were trained on Nvidia Quadro RTX 8000 GPUs. Training a \method{} on a single GPU takes around 20 hours on average.

\end{document}